\documentclass[letterpaper]{article} \usepackage{mystyle}
\usepackage{aaai2026}
\usepackage{times}  \usepackage{helvet}  \usepackage{courier}  \usepackage[hyphens]{url}  \usepackage{graphicx}    \usepackage{natbib}  \usepackage{caption}     \usepackage{algorithm}
\usepackage{algorithmic}
\usepackage{makecell}
\usepackage{newfloat}
\usepackage{listings}
\DeclareCaptionStyle{ruled}{labelfont=normalfont,labelsep=colon,strut=off} 
\floatstyle{ruled}
\newfloat{listing}{tb}{lst}{}
\floatname{listing}{Listing}
\title{
    \PRISM: Phase-enhanced Radial-based Image Signature Mapping\\
    framework for fingerprinting AI-generated images
}
\author {
Emanuele Ricco\textsuperscript{\rm 1},
    Elia Onofri\textsuperscript{\rm 2},
    Lorenzo Cima\textsuperscript{\rm 3},
    Stefano Cresci\textsuperscript{\rm 3},
    Roberto Di Pietro\textsuperscript{\rm 1}
}
\affiliations {
\textsuperscript{\rm 1} King-Abdullah University of Science and Technology (KAUST), CEMSE division, Thuwal, Saudi Arabia \\
    \textsuperscript{\rm 2} Institute for Applied Mathematics, National Research Council (IAC-CNR), Italy \\
    \textsuperscript{\rm 3} IIT-CNR, Pisa, Italy \\
    name.surname@kaust.edu.sa,
    name.surname@iit.cnr.it,
    elia.onofri@iac.cnr.it
}

\begin{document}

\maketitle

\begin{abstract}
A critical need has emerged for generative AI: attribution methods.
That is, solutions that can identify the model originating AI-generated content.
This feature, generally relevant in multimodal applications, is especially sensitive in commercial settings where users subscribe to paid proprietary services and expect guarantees about the source of the content they receive.
To address these issues, we introduce \PRISM, a scalable Phase-enhanced Radial-based Image Signature Mapping framework for fingerprinting AI-generated images.
\PRISM is based on a radial reduction of the discrete Fourier transform that leverages amplitude and phase information to capture model-specific signatures.
The output of the above process is subsequently clustered via linear discriminant analysis to achieve reliable model attribution in diverse settings, even if the model's internal details are inaccessible.
To support our work, we construct \PRISMDB, a novel dataset of 36,000 images generated by six text-to-image GAN- and diffusion-based models.
On this dataset, \PRISM  achieves an attribution \textit{accuracy} of 92.04\%.
We additionally evaluate our method on four benchmarks from the literature, reaching an average \textit{accuracy} of $81.60\%$.
Finally, we evaluate our methodology also in the binary task of detecting \textit{real} vs \textit{fake} images, achieving an average \textit{accuracy} of 88.41\%. We obtain our best result on \GENIMAGE with an \textit{accuracy} of 95.06\%, whereas the original benchmark achieved 82.20\%.
Our results demonstrate the effectiveness of frequency-domain fingerprinting for cross-architecture and cross-dataset model attribution, offering a viable solution for enforcing accountability and trust in generative AI systems.
\end{abstract}

\section{Introduction}\label{sec:introduction} 
In recent years, generative Artificial Intelligence (AI) has experienced an unprecedented surge, becoming a foundational pillar of modern computing. Powered by advances in deep learning and the surge in computing capabilities, these systems can generate fluent text, realistic images, synthetic speeches, and even coordinate cross-modal reasoning tasks with increasing human-like coherence~\cite{cao2023comprehensive}.
The rapid adoption of generative AI technologies in both industry and society reflects their transformative potential. Multimodal AI, in particular, unlocks new frontiers by seamlessly integrating vision, language, and audio understanding, enabling more natural and context-rich human-computer interactions~\cite{zhan2023multimodal}.
While the benefits are relevant and immediate, the fast-paced integration of generative AI raises important security and trust concerns that must be addressed with an adequate level of rigour and formalism, preserving applicability. 

\begin{figure}[t]
    \centering
    \includegraphics[width=.95\linewidth]{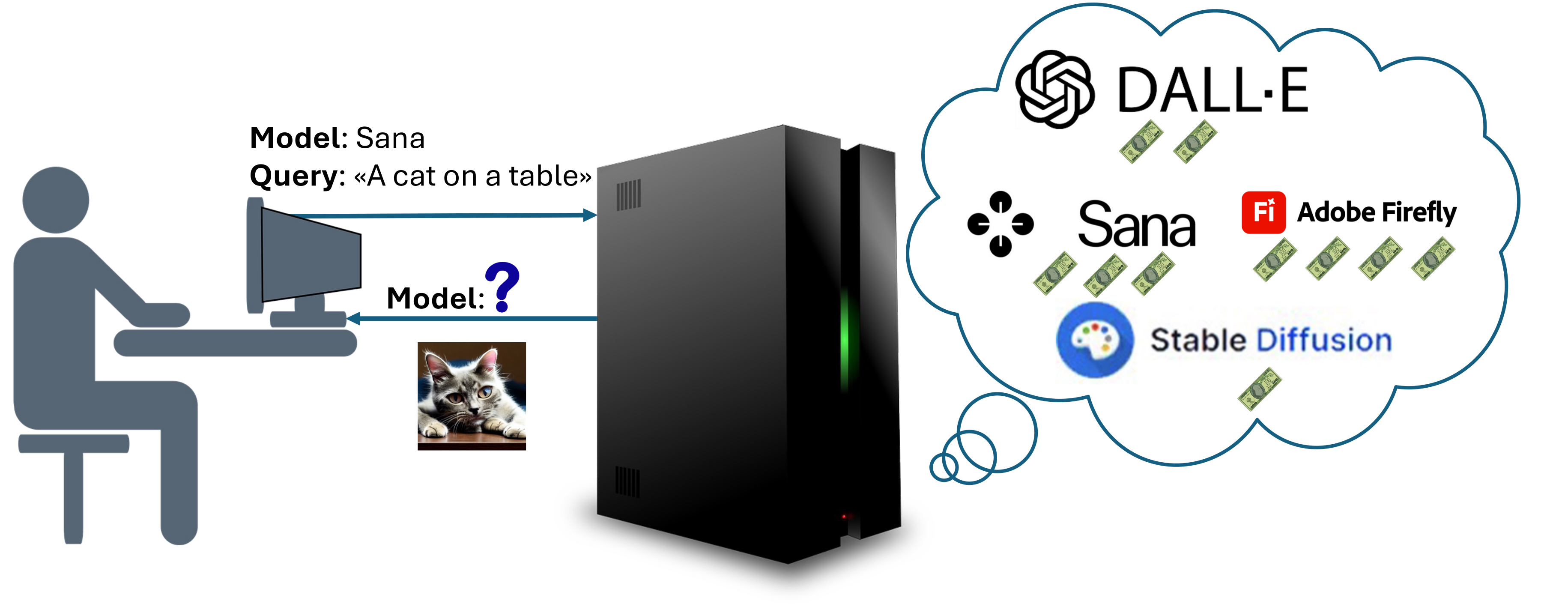}
    \caption{Problem overview: users want assurance that the responses they receive originate from the specific model they have subscribed to.}
    \label{fig:problem}
\end{figure}
As generative AI is increasingly deployed in real-world social contexts, understanding its strengths and vulnerabilities has become a societal necessity to ensure its safe use~\cite{pasupuleti2023cyber,cima2025contextualized}.
A fundamental step is understanding the distinct characteristics of the generated artefacts, as this bears important practical implications. For example, by systematically analysing how different model configurations --such as architecture, size, training data, and decoding strategies-- influence output features, it is possible to identify specific patterns that can serve as reliable signatures of the generated artifacts~\cite{wissmann2024whodunit,xu2025detecting}. In turn, this can enable model fingerprinting, surpassing the binary task of \textit{real} vs \textit{fake} detection~\cite{coccomini2024deepfake, sha2024zerofake, tan2024frequency}. Assessing the authenticity (\ie, fingerprinting) of the source model that generated an artefact is both theoretically and practically important.
Theoretically, since the black-box nature of large-scale generative models trades off performance for transparency~\cite{cao2023comprehensive}. Practically, because the need for output-level fingerprinting is growing, especially in scenarios where commercial (\ie, premium) APIs provide access to proprietary models \cite{ramesh2022dalle2}. Indeed, as shown in Figure~\ref{fig:problem}, end users may seek assurance that the generated artefacts they receive actually originate from the model they subscribed to, rather than from a---likely cheap---different one~\cite{wissmann2024whodunit}. 

In this work, we move beyond the traditional binary distinction between human- and AI-generated content by tackling the challenging task of fingerprinting multiple models in the context of image generation \cite{wang2023did}. Specifically, given an image and a set of possible sources (either natural images or AI-generated), we aim at identifying the model that generated the image.

To this end, we introduce \PRISM, a novel fingerprinting framework leveraging both the amplitude and phase components of the radial-reduced Discrete Fourier Transform (DFT) to capture subtle, yet informative, model-specific characteristics.
Model attribution is then obtained with a supervised clustering step via Linear Discriminant Analysis (LDA). To validate \PRISM, we create and publish a reference dataset of 36,000 images generated by six models, based on either GAN or diffusion architectures, from 40 distinct textual prompts. We evaluate \PRISM on this dataset and on four additional reference datasets, achieving excellent and generalizable results. Our results underscore the potential of frequency-domain fingerprinting as a robust solution for model attribution, an essential step toward ensuring accountability, trust, and provenance in real-world deployments of generative AI systems.

\paragraph{Contributions}
\label{sub:contribution}
Our contributions related to AI-generated image fingerprinting can be summarised as follows:
\begin{itemize} 
    \item We propose \PRISM, a model-agnostic LDA-based framework that, exploiting radially-reduced frequency-domain features, accurately fingerprints AI-generated images to identify their source models.
    \item We generate and release \PRISMDB, a novel dataset of $36,000$ images generated by six text-to-image (T2I) models, including both open-source and closed models.
    \item We evaluate our methodology across \PRISMDB and four publicly available benchmarks, achieving attribution accuracies of 92.0\% on our dataset and between 76.3\% and 88.5\% on the existing ones.
    \item We further evaluate \PRISM on the binary task of detecting \textit{real} vs \textit{fake} images, achieving an average accuracy of 88.41\% and demonstrating its suitability in this task.
\item We release both data and code,\footnote{A reduced version is available at: \url{https://shorturl.at/BBl0m}} 
    to allow full reproducibility of our results.
\end{itemize}

\section{Related Work}
\label{sec:related}
In the last few years, the content generated by AI has led to the flowering of many techniques to distinguish real from artifact data, due to the introduction of many techniques for image generation. The theme of model-specific fingerprinting is crucial for many reasons, such as intellectual property rights to ensure users comply with their legal licenses, as investigated in \cite{xu2024instructional}.

The rapid advancement and widespread deployment of LLMs highlight the growing need for robust model attribution techniques, which are essential for identifying the source model and mitigating adversarial misuse. A commonly explored strategy involves watermarking generated outputs, \eg embedding patterns through neutral token sequences~\cite{kirchenbauer2023watermark} or leveraging a powerful LLM, such as ChatGPT, to watermark outputs produced by other models~\cite{zhong2024watermarking}.
More recently, attention was also drawn to model fingerprinting not requiring the need for explicit watermarking.

In the textual domain, various approaches have emerged: to cite a few, in~\cite{zeng2024huref} the authors leverage architectural patterns in transformers to identify three unique terms that distinguish base models, while~\cite{alhazbi2025llms} propose a method based on timing intervals between token generations.
Other efforts explore implicit markers~\cite{wanli2025imf}, the scalability of fingerprinting techniques~\cite{nasery2025scalable}, and the robustness against model merging~\cite{yamabe2024mergeprint}. However, these fingerprinting approaches are focused exclusively on text-based LLMs.

Attributing AI-generated images to their source models has been a concern since the early days of GANs.
Initial work showed that minor differences in architecture could leave identifiable output fingerprints~\cite{yu2019attributing}, prompting studies on how model structure shapes these patterns~\cite{marra2019gans}.
A common approach involves embedding watermarks in training data to test their robustness against output perturbations~\cite{yu2021artificial}.
The rise of diffusion models has intensified interest in tracking image provenance~\cite{ho2020denoising}.
Techniques include injecting frequency-domain patterns into noise inputs~\cite{wen2023tree} or embedding signatures directly into model weights~\cite{fernandez2023stable}.
Other strategies use latent fingerprinting or user-specific digital signatures for accountability~\cite{nie2023attributing,kim2024wouaf}.

However, these approaches rely on low-level access to the model's architecture or parameters—an assumption that does not hold for commercial platforms such as \texttt{Midjourney} or \DALLE, where internal details remain undisclosed. To overcome this limitation, we introduce a scalable, model-agnostic fingerprinting framework that operates without requiring access to proprietary internals. Unlike previous work that focuses mainly on distinguishing between different versions of the same model~\cite{wissmann2024whodunit}, our method tackles a distinct and novel task by attributing outputs to entirely different model families. A detailed comparison with other model-agnostic techniques is available in Appendix~\ref{app:comparative_charts}, Table~\ref{tab:fingerprinting-comparison}.

\section{Datasets for Model Attribution}
\label{sec:materials}
We consider an image as its split in RGB channels, namely a triplet of matrices in $\{0, \dots, 255\}^{n_y \times n_x}$, where $n_x$ and $n_y$ represent the bit-size of the image, obtained as the actual size multiplied by the image resolution. Formally, the set of all possible images is given by $\III = \bigcup_{n_x, n_y\in\NN}(\NN_{256}^{n_y\times n_x})^3$. Then, let $\XXX$ be a set of $N$ images generated by a set $\MMM$ of $M$ models. We define a dataset for the attribution task as the couple $\DDD = (\XXX, \YYY)$, where $y_i \in \YYY$ is the source of the image $x_i \in \XXX$, with $x_i\in\III$, $y_i \in \MMM$.

Many datasets with generated images are available in the literature. However, these are often built for purposes such as fake image detection, visual quality evaluation, and identity preservation, and are therefore not suitable to be directly used for the model attribution task. For this reason, we first created a dataset specifically designed for this task. 

\begin{figure}[t]
    \centering
\scalebox{.7}{

\begin{tikzpicture}[
        blbx/.style = {draw=black, inner sep=0pt, align=center},
    ]

    \node[blbx, inner sep=.4em] (prompt) at (-0, 0) {
        \footnotesize \textit{``A scene from}\\[-.2em]
        \footnotesize \textit{War and Peace}\\[-.2em]
        \footnotesize \textit{reinterpreted in a}\\[-.2em]
        \footnotesize \textit{futuristic landscape}\\[-.2em]
        \footnotesize \textit{with robots instead}\\[-.2em]
        \footnotesize \textit{of humans.''}
    };

    \node[blbx, label={[label distance=-.11em]90:\footnotesize \VQGAN}] (1a) at (3, 1.4) {\includegraphics[width=2cm]{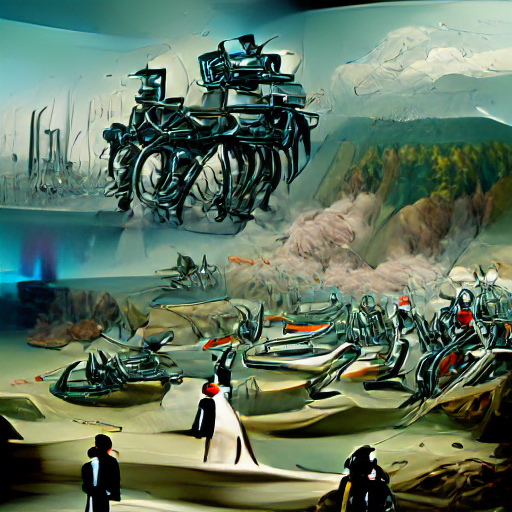}};
    \node[blbx, label={below:\footnotesize \PIXART}] (1b) at (3, -1.4) {\includegraphics[width=2cm]{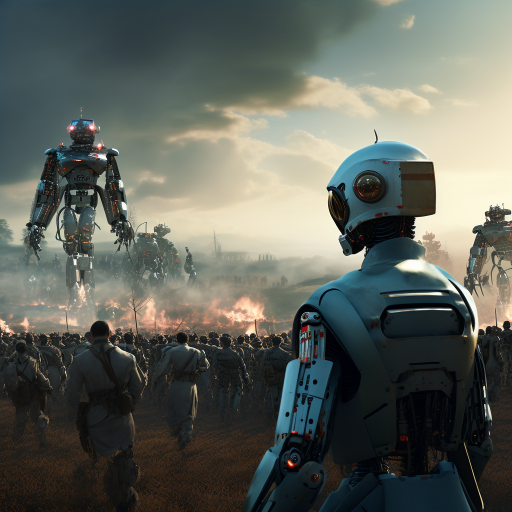}};
    
    \node[blbx, label={above:\footnotesize \DALLE}] (2a) at (5.2, 1.4) {\includegraphics[width=2cm]{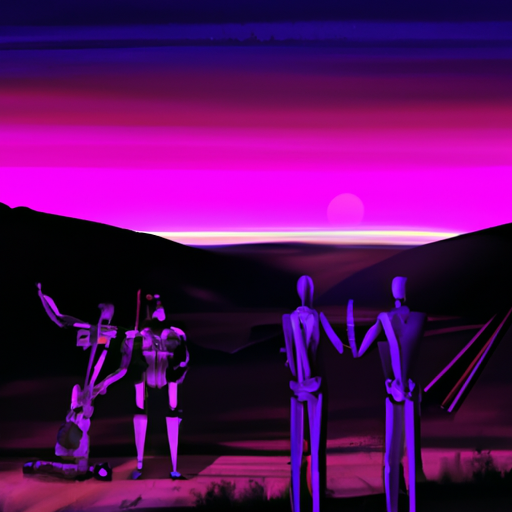}};
    \node[blbx, label={below:\footnotesize \SANA}] (2b) at (5.2, -1.4) {\includegraphics[width=2cm]{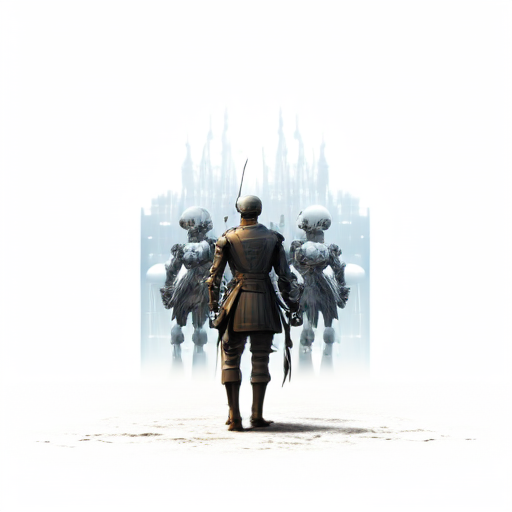}};
    
    \node[blbx, label={above:\footnotesize \FUSEDREAM}] (3a) at (7.4, 1.4) {\includegraphics[width=2cm]{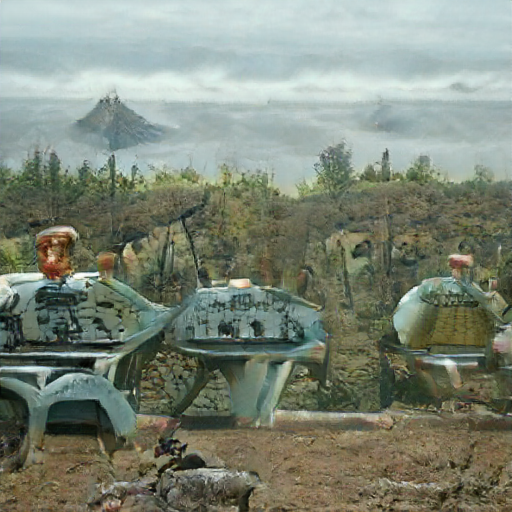}};
    \node[blbx, label={below:\footnotesize \SBnv}] (3b) at (7.4, -1.4) {\includegraphics[width=2cm]{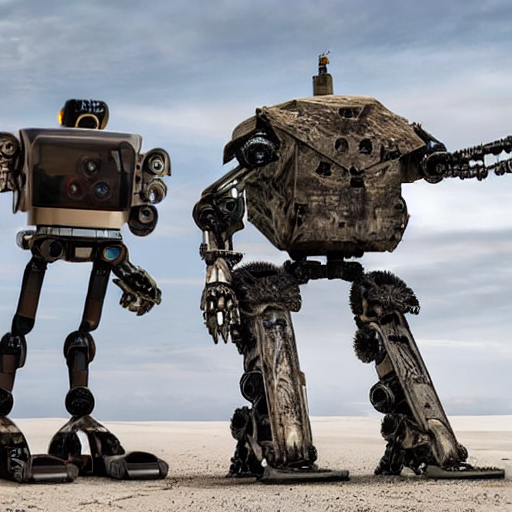}};

\end{tikzpicture} }
    \caption{
        Example of images generated by the six different models forming \PRISMDB with the same prompt.
    }
    \label{fig:dataset}
\end{figure}

\subsection{\PRISMDB}
\label{sec:our-dataset}
We built \PRISMDB to serve as a reference benchmark for the model attribution task
\footnote{A reduced version is available at: \url{https://shorturl.at/BBl0m}}. 
It comprises 36,000 $512\times512$ PNG images generated using six different T2I models selected based on the availability of open-access white papers. This ensures transparency, reproducibility, and the possibility of future extensions. Figure \ref{fig:dataset} shows examples of the images generated by the different models for the same prompt. Appendix Table~\ref{tab:our-dataset} thoroughly reports the characteristics of each model included in the dataset. In particular, we considered the image resolution capabilities, underlying architecture (\eg, GANs~\cite{reed2016generative}, Transformers~\cite{ding2021cogview,yu2022scaling}, Diffusion Models~\cite{ho2020denoising}, CLIP~\cite{radford2021learning}), and openness of their implementations.
For instance, \SB is fully open (code, architecture, and weights available under a permissive license), while models such as \VQGAN and \FUSEDREAM are only partially open due to dependencies on proprietary components like CLIP.
Conversely, \PIXART and \SANA are both released on GitHub under MIT and Apache 2.0 licenses respectively, yet being partially open as it is not possible to directly download the pre-trained weights.
Finally, \DALLE is the only closed model used, being accessible through paid APIs only.

To construct the dataset, we crafted a set $\PPP$ of 40 prompts (see Appendix~\ref{app:labels}) grouped into two categories: 20 short ($p_1, \dots p_{20}$) and 20 long ($p_{21}, \dots, P_{40}$).
Short prompts $p_i$ follow a simple [subject-verb-complement] syntax, while corresponding long prompts $p_{i+20}$ introduce additional contextual complexity over the short ones.
Each prompt was used to generate 150 images per model, resulting in 6,000 images per model.
Prompts were treated as auxiliary classes for dataset split purposes, with the aim of learning discriminative features ascribable to the generation source only while being prompt-agnostic. 

We also evaluated a carefully crafted \textit{average} split in training vs.\ test set (80:20), so that short-long couples are preserved, \ie, namely, ($p_i, p_{i+20}$) are always paired in the same set, fostering the resulting model foresee generalisation to previously unseen prompts.
Here, ``average'' refers to the split being cherry-picked to obtain as-average-as-possible results w.r.t.\ possible splits, instead of taking the one maximising, say, \textit{accuracy} (see later Section~\ref{sec:results}).

\subsection{Literature Datasets} \label{ssec:other-dataset}

To benchmark our approach and ensure comparability with prior work, we applied our methodology to several publicly available datasets. In particular, we selected four open-access datasets presenting published white-papers on their construction, namely: Generated Faces in the Wild (\GFW)~\cite{borji2022generated}, \DEEPGUARD~\cite{namani2025deepguard}, \SUSY~\cite{bernabeu2024present}, and \GENIMAGE~\cite{zhu2023genimage} (for this latter one, we used its tiny version instead{\footnote{\url{https://www.kaggle.com/datasets/yangsangtai/tiny-genimage}}}).

Details and statistics of each dataset are summarised in Appendix~\ref{app:comparative_charts} within Table~\ref{tab:different-dataset}, including image sources, format, resolution (being mostly varying across and within datasets), dataset size, and reported classes (typically \textit{real} and \textit{fake}). Along with real images (taken from \COCO or unpublished sources), notably a total of 16 different AI sources are present, namely:
\texttt{DALL-E} (v2, v3),
\texttt{Midjourney} (original, tti, img),
\SBnv (v1.4, v1.5, v1.x, v3.0),
\texttt{GLIDE},
\texttt{IMAGEN},
\texttt{SDXL}
\texttt{ABM},
\texttt{BiGGAN},
\texttt{VQDM},
\texttt{Wukong}.
It is also worth noticing that both \DEEPGUARD and \GENIMAGE organise images as couples, where \textit{fake} images are generated based on \textit{real} ones.

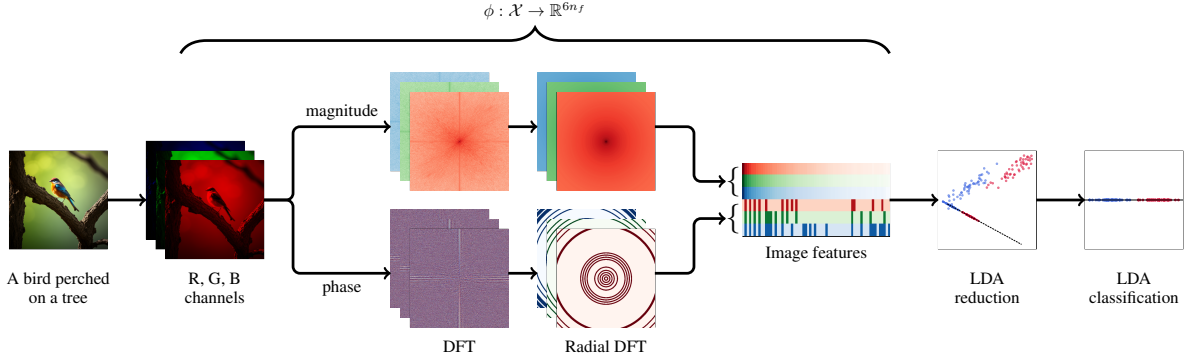
\begin{figure*}[!th]
    \centering
    \scalebox{.65}{{\begin{tikzpicture}[
        overbrace/.style={ultra thick, decoration={brace, amplitude=15pt}, decorate},
        blbx/.style = {draw=black, inner sep=0pt},
        bracenode/.style={pos=0.5,anchor=north,yshift=1.2cm},
    ]

    \node[align=center] at (-4, -1.8) {A bird perched\\ on a tree};
    \node[align=center] at (-.8, -1.8) {R, G, B\\ channels};
    \node[align=center] at (4.2, -3) {DFT};
    \node[align=center] at (7.2, -3) {Radial DFT};
    \node[align=center] at (11.5, -1.1) {Image features};
    \node[align=center] at (15, -1.8) {LDA\\reduction};
    \node[align=center] at (18, -1.8) {LDA\\classification};
    
    \node[blbx] (0) at (-4, 0) {\includegraphics[width=2cm, height=2cm]{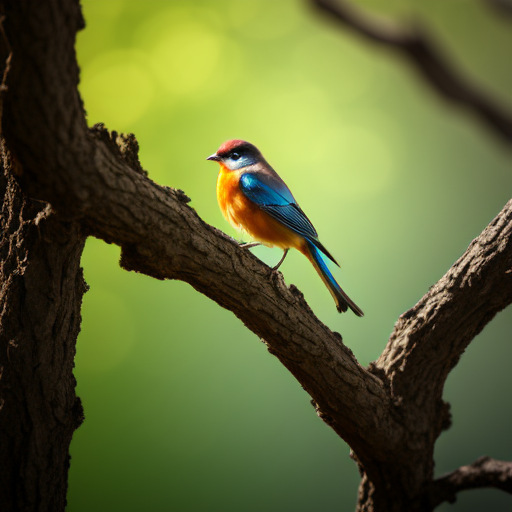}};
    
    \node[blbx] at (-1.2, .2) {\includegraphics[width=2cm, height=2cm]{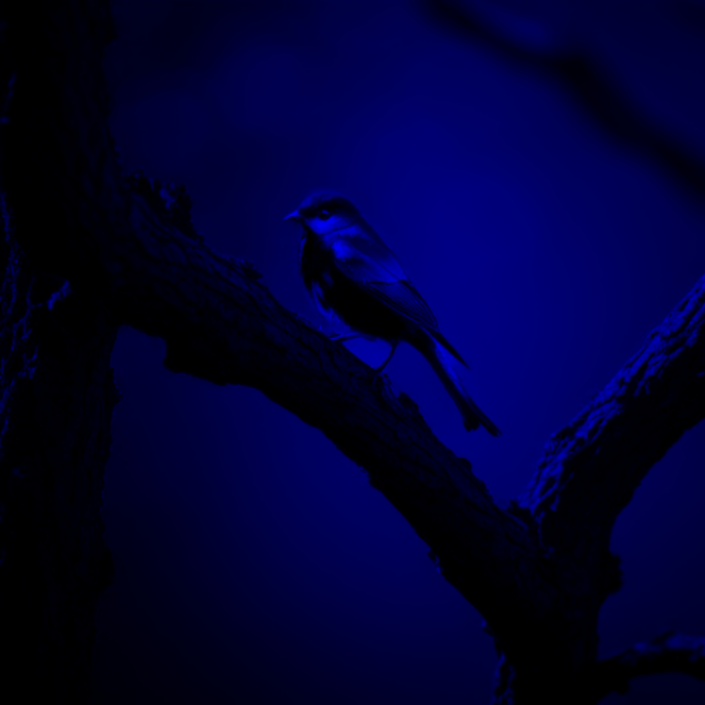}};
    \node[blbx] at (-1, 0) {\includegraphics[width=2cm, height=2cm]{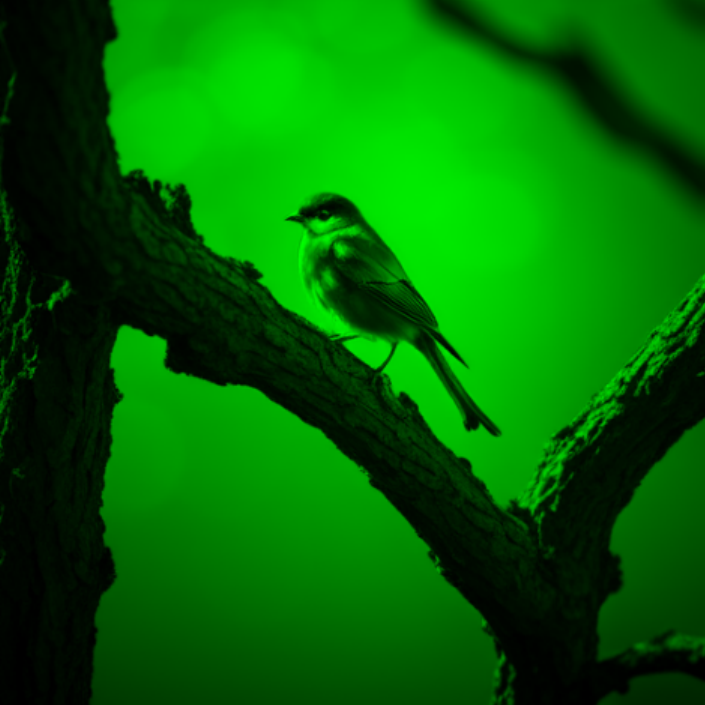}};
    \node[blbx] at (-.8, -.2) {\includegraphics[width=2cm, height=2cm]{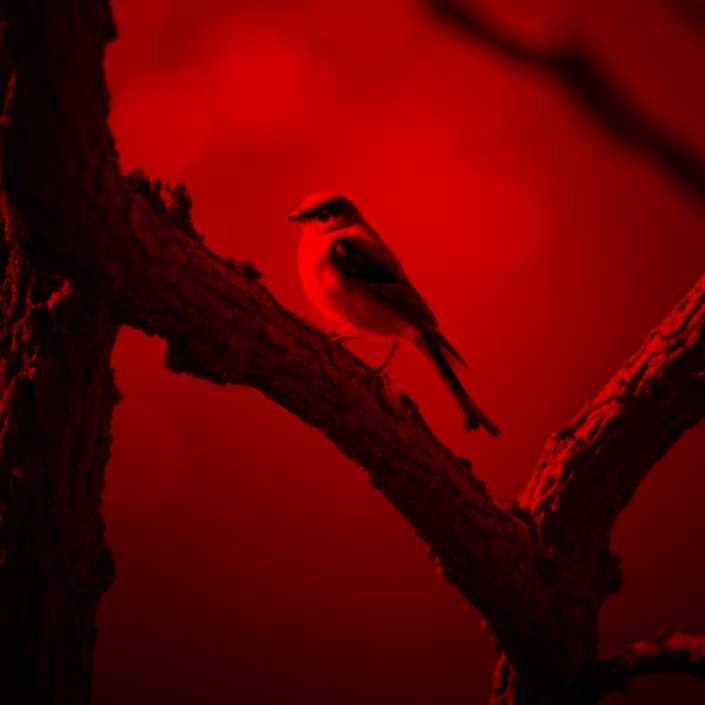}};

    \node[blbx] at (3.8, 1.6) {\includegraphics[width=2cm, height=2cm]{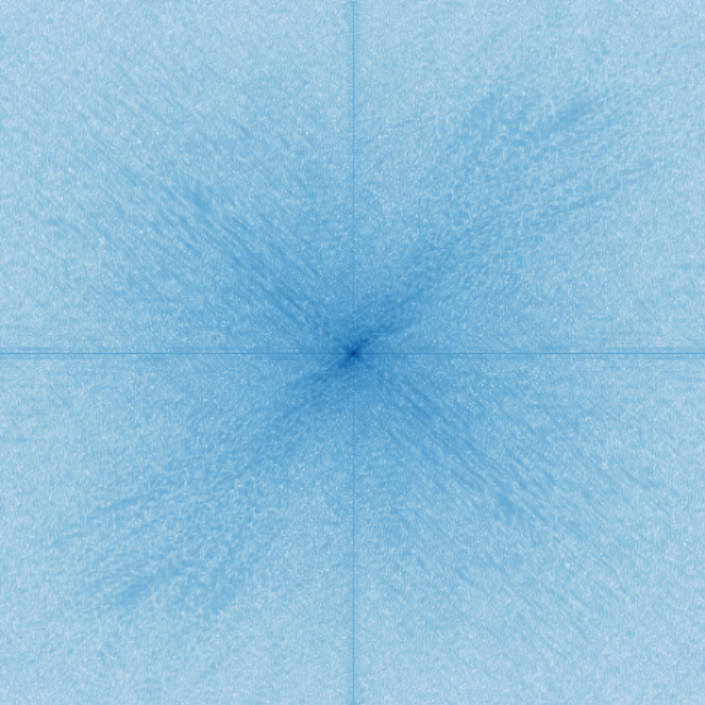}};
    \node[blbx] at (4.0, 1.4) {\includegraphics[width=2cm, height=2cm]{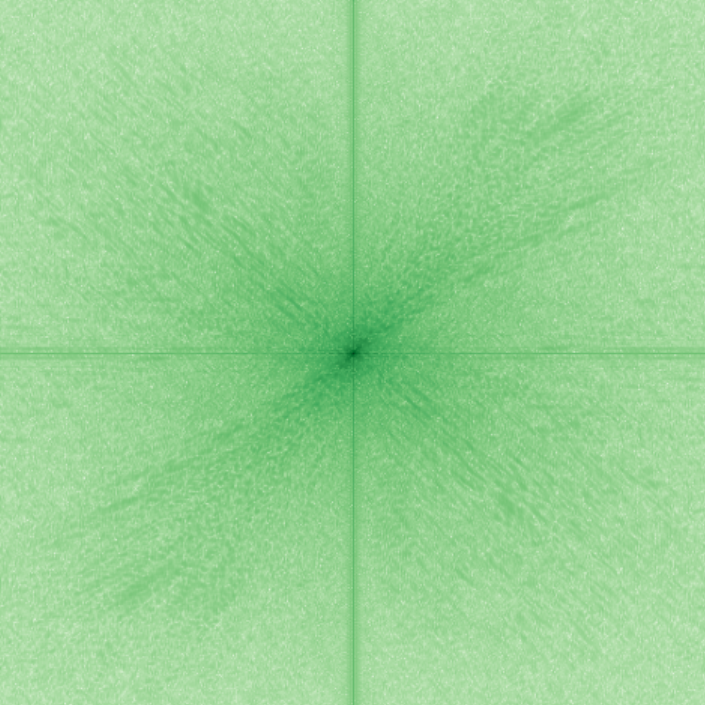}};
    \node[blbx] at (4.2, 1.2) {\includegraphics[width=2cm, height=2cm]{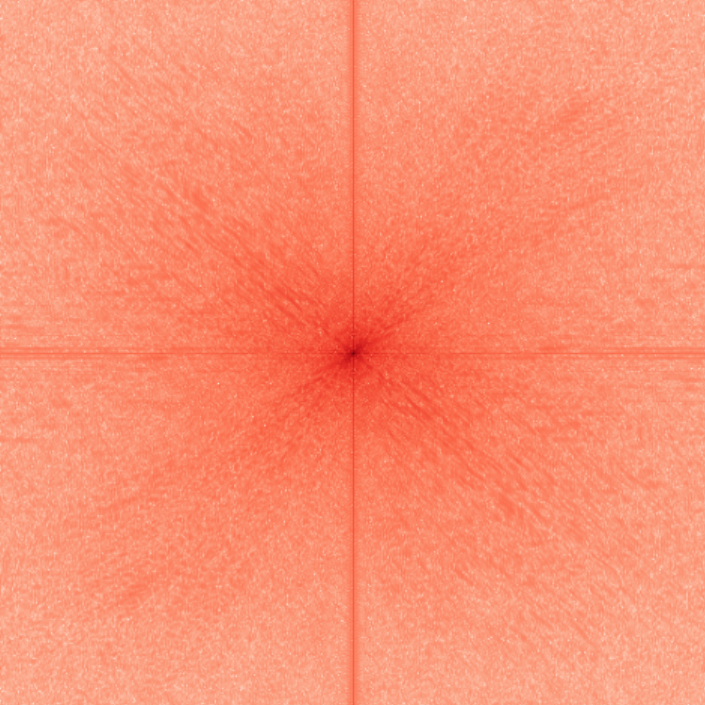}};
    
    \node[blbx] at (3.8, -1.2) {\includegraphics[width=2cm, height=2cm]{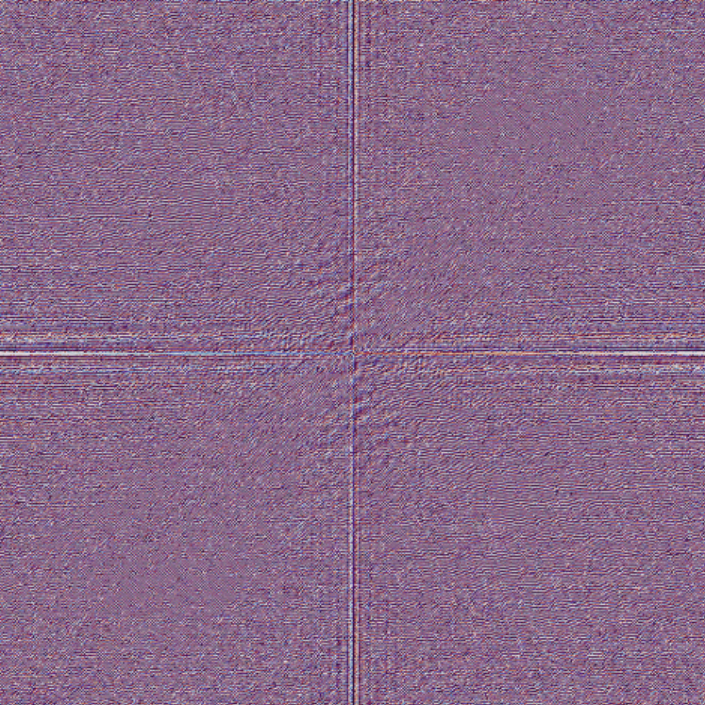}};
    \node[blbx] at (4.0, -1.4) {\includegraphics[width=2cm, height=2cm]{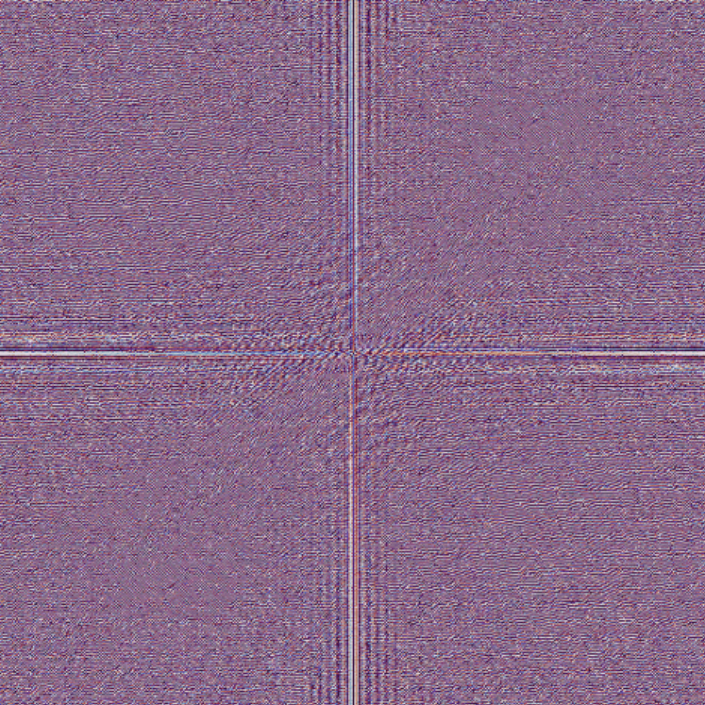}};
    \node[blbx] at (4.2, -1.6) {\includegraphics[width=2cm, height=2cm]{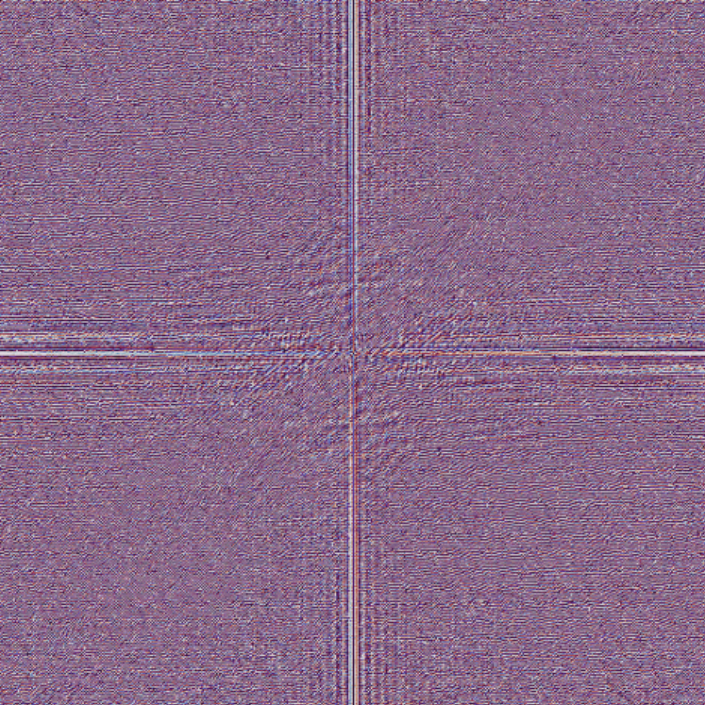}};

    \node[blbx] at (6.8, 1.6) {\includegraphics[width=2cm, height=2cm]{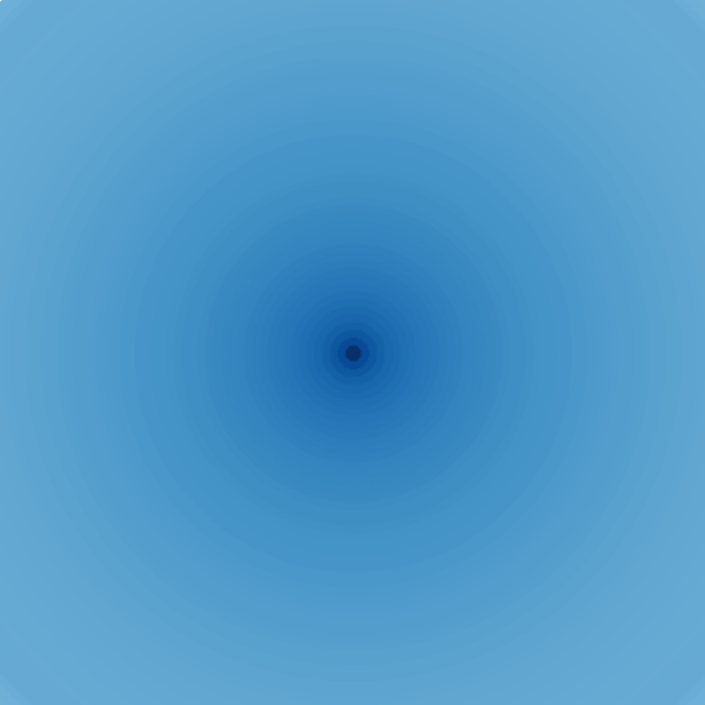}};
    \node[blbx] at (7.0, 1.4) {\includegraphics[width=2cm, height=2cm]{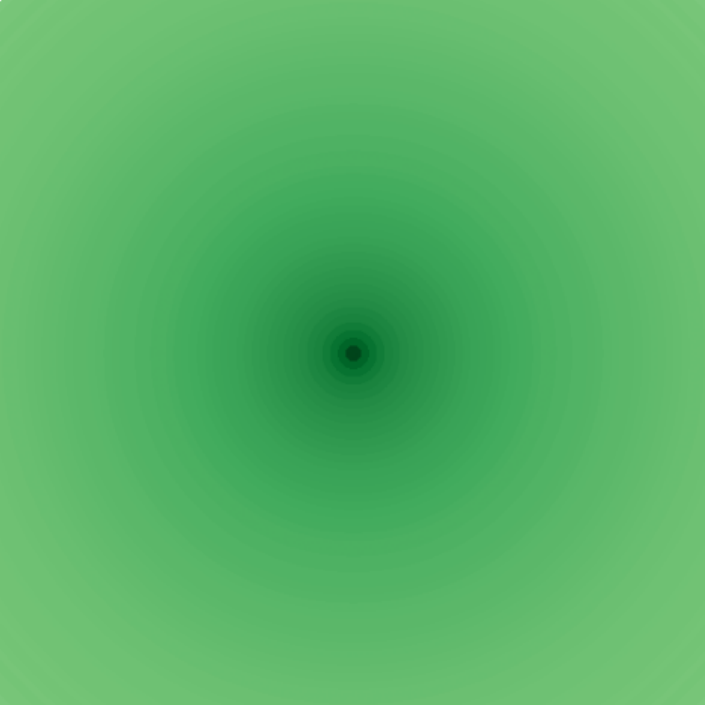}};
    \node[blbx] at (7.2, 1.2) {\includegraphics[width=2cm, height=2cm]{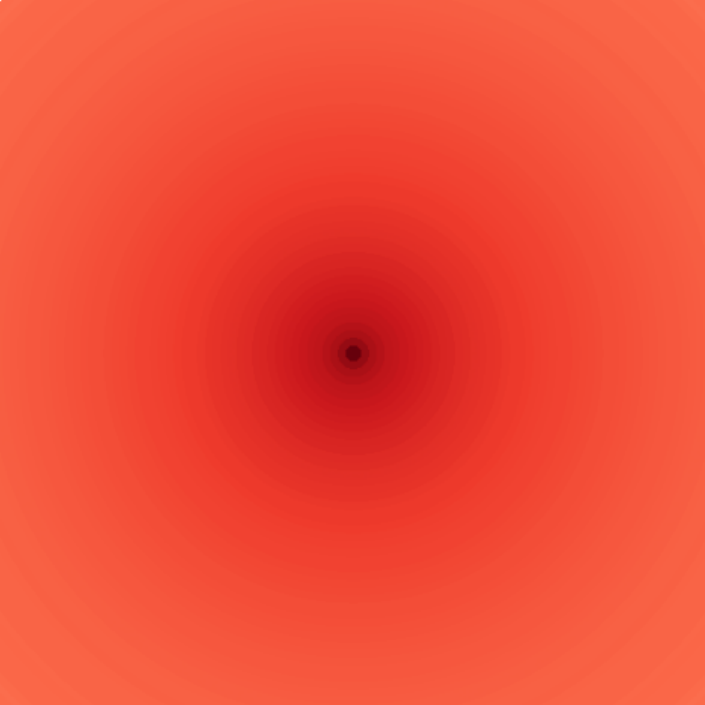}};
    
    \node[blbx] at (6.8, -1.2) {\includegraphics[width=2cm, height=2cm]{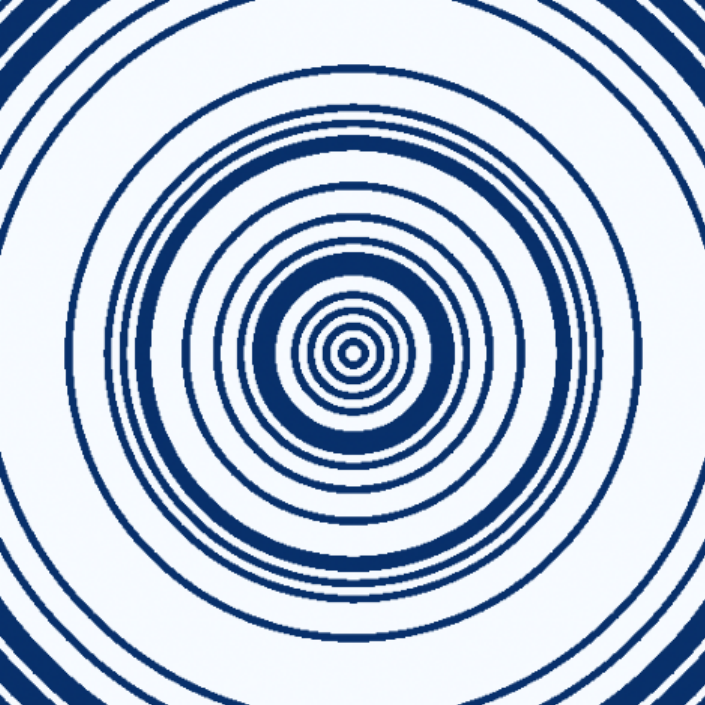}};
    \node[blbx] (0) at (7.0, -1.4) {\includegraphics[width=2cm, height=2cm]{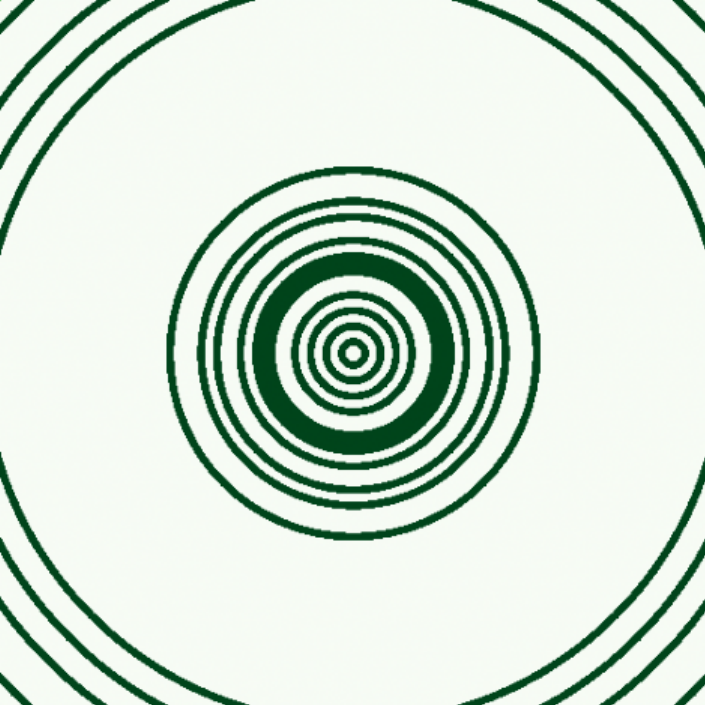}};
    \node[blbx] (0) at (7.2, -1.6) {\includegraphics[width=2cm, height=2cm]{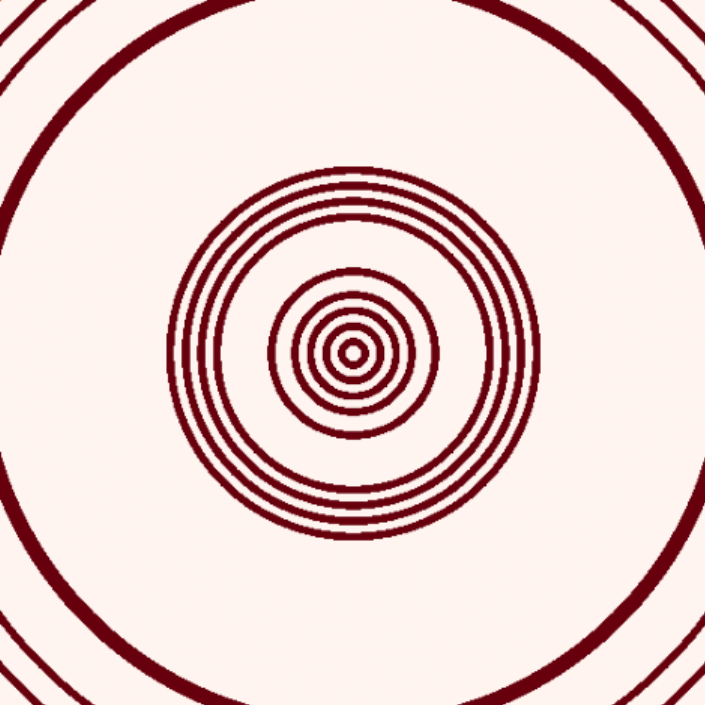}};
    
    \node[blbx] (0) at (11.5, 0) {\includegraphics[width=3cm]{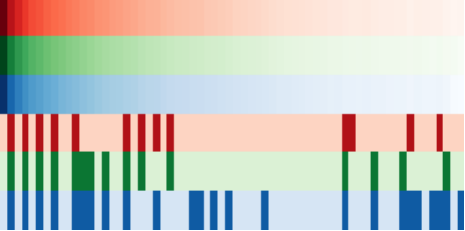}};

    \draw[->, ultra thick] (-3, 0) -- (-2.2, 0);
    \draw[->, ultra thick, rounded corners] (0.2, 0) -| (.8, 1.5) --node[above] {magnitude} (2.8, 1.5);
    \draw[->, ultra thick, rounded corners] (0.2, 0) -| (.8, -1.5) --node[below]{phase} (2.8, -1.5);
    \draw[->, ultra thick] (5.2, 1.5) -- (5.8, 1.5);
    \draw[->, ultra thick] (5.2, -1.5) -- (5.8, -1.5);
    \draw[->, ultra thick, rounded corners] (8.2, 1.5) -| (9.0, .38) -- (9.65, .38);
    \draw[->, ultra thick, rounded corners] (8.2, -1.5) -| (9.0, -.38) -- (9.65, -.38);

    \node[] at (9.8, +.38) {\huge \{};
    \node[] at (9.8, -.38) {\huge \{};
    
    \draw [overbrace] (-1.5,3) -- ++(14.5,0) node [bracenode] {$\phi:\XXX \to \RR^{6n_f}$};

    \node[blbx] at (15, 0) {\includegraphics[height=2cm, width=2cm]{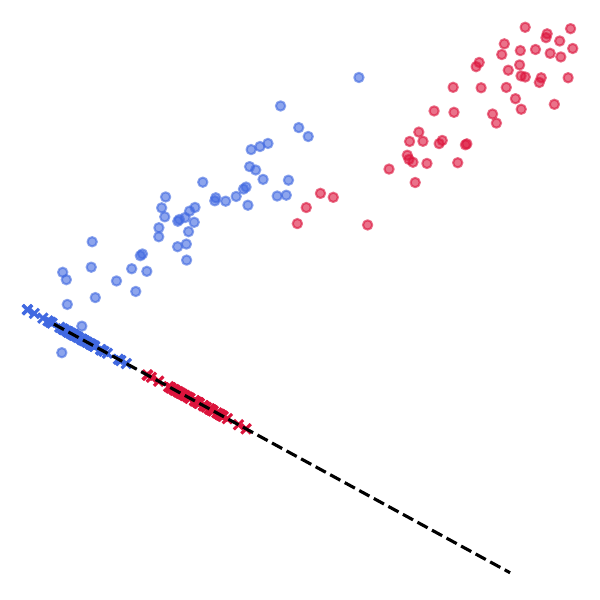}};

    \node[blbx] at (18, 0) {\includegraphics[height=2cm, width=2cm]{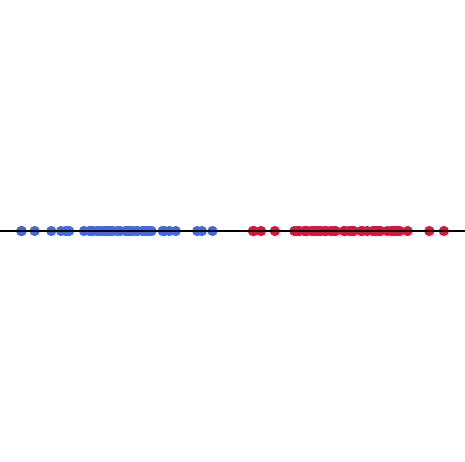}};

    \draw[->, ultra thick] (13, 0) -- (14,0);
    \draw[->, ultra thick] (16, 0) -- (17,0);
    
\end{tikzpicture} }}
    \caption{
        Overview of \PRISM methodology.
        Given an image, feature extraction ($\phi$) is applied prior to dimensionality reduction and classification with LDA.
    }
    \label{fig:pipeline}
\end{figure*}

\section{Methodology} \label{sec:methodology}
We now present the fundamental components of our proposed methodology, composed of an image magnitude-and-phase feature extraction through a radially-reduced DFT of the RGB channels, and the usage of LDA both as a dimensionality reduction and classification algorithm (see Figure \ref{fig:pipeline}). 
DFT, in fact, has been previously employed in the literature for frequency-based deepfake detection~\cite{karageorgiou2025any} and artefact analysis in generated images~\cite{tan2024frequency, li2025improving}, with the latter also incorporating phase information with the amplitude to enhance detection. 

\subsection{Feature Extraction and Analysis}
\label{ssec:rFFT}
Formally, we can define the ``model attribution'' problem as the task of training a discriminant function:
\begin{equation}\label{eq:a}
    \begin{matrix}
        a : & \XXX & \to      & \MMM\\
            & x     & \mapsto & m
    \end{matrix}\ ,
\end{equation}
mapping a given image $x$ to the corresponding source $m$.
It is worth noting that \eqref{eq:a} is a natural generalisation of the more renowned task of ``fake detection'' which reduces the sources space $\MMM$ to the binary classification of \textit{real} vs \textit{fake}, and hence training a simpler function 
\(
a_f : \XXX \to \{0, 1\}\ .
\)

Due to the nature of our analysis, we decided to build our dataset in the lossless format of Portable Network Graphics (PNG), representing $512\times512$ images directly as 3(4) $512\times512$ matrices for the RGB(A) channels --the transparency layer A is constant 1 in our experiments and hence omitted.
Moreover, we incorporated datasets from the literature containing images in the JPEG and WebP lossy formats.

Each colour channel is subsequently transformed into the frequency domain via a two-dimensional DFT, followed by frequency centralisation (see Appendix~\ref{app:DFT} for further reference).
We then compute the corresponding radial profile --referred to as radial DFT (rDFT)-- by averaging the spectral log-magnitude $\overline M_{u, v}$ and phase $\overline \Phi_{u, v}$ across $n_r = 64$ evenly spaced annular region $\BBB_1, \dots, \BBB_{n_r}$ spanning the frequency space.
More in detail, for each bin $i = 1, \dots, n_r$, we compute the mean of the log-magnitude values and the cosine-transform of the circular mean of the phase angles as:
\begin{align}
M_i &= \frac{1}{|\BBB_i|} \sum_{(u,v) \in \mathcal{B}_i} \overline{M}_{u, v},\label{eq:M_i} \\
	\Phi_i &= \cos\left( \arg\left( \frac{1}{|\BBB_i|} \sum_{(u,v) \in \mathcal{B}_i} e^{i \overline{\Phi}_{u,v}} \right)\right)\ ,\label{eq:Phi_i}
\end{align}
where the cosine-transform is used to identify $-\pi$ and $\pi$ angles without any potential loss of information (since the matrix is Hermitian, the aggregated phase collapses on $\{0, \pi\}$).
This yields a compact radial representation of both magnitude and phase spectra, while also standardising the approach w.r.t.\ the possible image dimensions $n_x$, $n_y$ and fostering the resilience against different lossy formats.

\begin{figure}
    \centering
    \includegraphics[width=.9\linewidth]{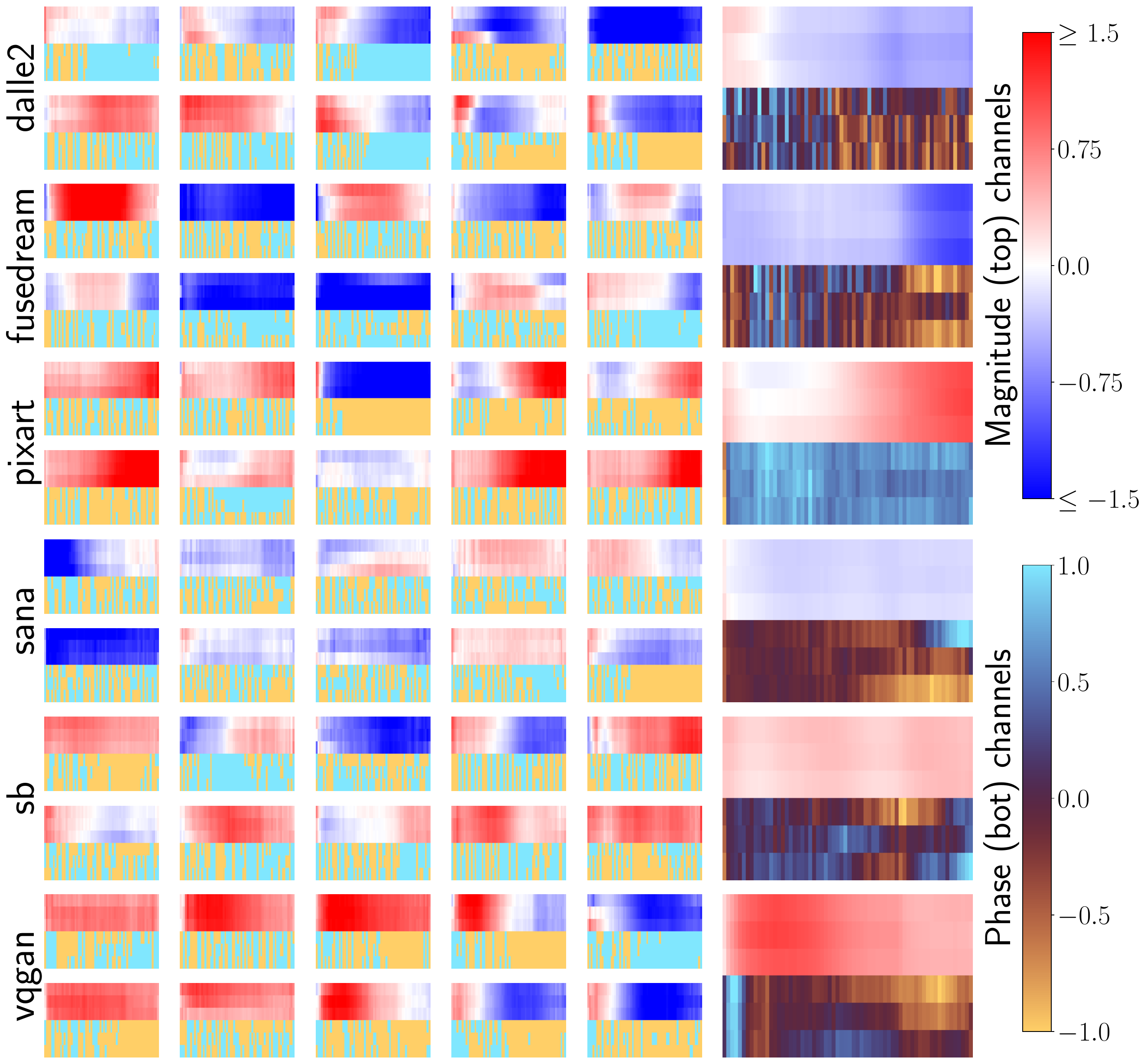}
    \caption{
        Examples of model fingerprints.
        Each small box represents the features extracted from a single generated image, with the upper half displaying the Fourier log magnitude (normalised) and the lower half showing the corresponding cosine-transformed phase.
        Each rightmost (big) box illustrates the average spectrum for the corresponding AI model, built as the median magnitude and the average phase spectra within all the model samples.
    }
    \label{fig:example-features-extraction}
\end{figure}

The aggregated descriptors for each colour channel are finally concatenated into a feature vector $\mathbf{f} \in \RR^{6n_r}$, later used in downstream analysis tasks.
In what follows, we identify with $\phi:\XXX \to \RR^{6n_r}$ this procedure, as also depicted in Figure~\ref{fig:pipeline}.
Figure~\ref{fig:example-features-extraction} reports a few examples of such vectors for the various models considered in our analysis, along with the average value of all images per model.

\subsection{Model Attribution Pipeline}

Let us consider our dataset $\DDD = \{\XXX, \YYY\}$ and let us denote with $\FFF = \{\mathbf{f}_i\}_{i=1}^{N}$ the set of feature vectors $\mathbf{f}_i \in \RR^{6n_f}$ obtained as $\mathbf{f}_i = \phi(x_i) \in \XXX$ through the rDFT feature extraction procedure described in Section~\ref{ssec:rFFT} and normalised.
As it is common in the literature, designing a discriminant function $a$ (cf.\ Eq.\ \eqref{eq:a}) mainly involves two distinct tasks:
\begin{itemize}
    \item choosing a suitable dimensionality reduction $\rho : \RR^{6n_f} \to \RR^{n_e}$, for some reduced number of features $n_e \leq 6n_f$, to project the data into a more tractable embedding space;
    \item identifying a classification strategy $\kappa : \RR^{n_e} \to C$, where $C$ denotes a set of $k \geq M$ clusters.
\end{itemize}
Particularly, when $C = \MMM$, then it is straightforward to compose $\kappa$, and $\rho$ with the feature extractor $\phi$ to create a valid discriminant function $a$:
\begin{equation}\label{eq:unsupervised-a}
    \begin{split}
        a =&\ \kappa \circ \rho \circ \phi \\
        a \,\,:\,&\ \XXX \to \RR^{6n_f} \to \RR^{n_e} \to \MMM
    \end{split}
\end{equation}

Amongst the many dimensionality reduction techniques available in the literature, we focus in particular on LDA~\cite{zhao2024linear}.
In fact, unlike unsupervised techniques such as, \eg, PCA and $t$-SNE, LDA is a supervised method that explicitly incorporates class label information to optimise separability.
In detail, LDA seeks a linear projection maximising class separability by the optimisation of the ratio between inter-class and intra-class variance.
This results in a $M-1$ dimensional embedding where clusters (now interpreted as known classes) are optimally separated, naturally aligning dimensionality reduction with classification.
In fact, along with the inherent dimensionality reduction offered by the embedding, LDA also admits a generative probabilistic interpretation, where each class is modelled as a Gaussian distribution with shared covariance; in this predictive formulation, it can be directly used as a classifier without requiring an additional clustering step (actually directly merging $\rho$ and $\kappa$ contributions).

\subsection{Evaluating the Results}
\label{sec:scores}

Upon defining the attribution pipeline via the supervised discriminant function \eqref{eq:unsupervised-a}, the final step consists of selecting appropriate performance indicators.
Given the semi-supervised nature of the proposed approach, we adopt four classical classification metrics (\textit{accuracy}, \textit{precision}, \textit{recall} and \fscore), all computed in their weighted form to account for the multi-class nature of the problem.

To ensure a robust evaluation and mitigate randomness in dataset splits, we computed all scores over $N_s$ randomly sampled 80:20 train-test partitions, while preserving the different prompts (\ie 16 vs.\ 4 couples of prompts per split), hence guaranteeing prompt generalisation.
This repeated evaluation serves a dual purpose: it provides an aggregated view of the typical performance in different close-to-random test conditions and supports the assessment of the method's generalisation capabilities to unseen data.

In addition, we designate one reference split --selected as the one whose performance is closest to the mean across all splits-- as the ``average'' split.
This allows for a fair, standardised comparison among different methods or configurations, and provides a concrete instance on which to present detailed performance insights.

\section{Results} \label{sec:results}
In what follows, we firstly present the averaged results obtained from $N_s = 1000$ different \PRISMDB random splits and we briefly analyse the relevance of the extracted features.
Then, we present the results for the average split, selected as the most representative within the $N_s$ we generated. 
Finally, we report the results obtained on the benchmark reference datasets introduced in Section~\ref{ssec:other-dataset}.

We run all the experiments under Python 3.12.9 with scikit-learn 1.6.1 on an Intel Xeon W7-3465X workstation equipped with 28/56 CPU threads (4.8 GHz) and 1 TB RAM, and repeated the tests for the sake of comparison on a MacBook Pro M3 Max Laptop, equipped with 16 CPU threads (4.05/2.75 GHz) and 64GB RAM.
The Ubuntu 24.04 workstation swiftly handled the feature extraction on \PRISMDB, yielding 384 features for each of the 36K images in $\sim$1 h (110.59 MB memory usage).
The radial reduction proved to be necessary, both for standardising the feature size (regardless the image original dimension) and for computational viability. Indeed, the extraction with the sole DFT yielded a dataset of 452.98 GB with 1,572,864 features per image ($\sim$20 min).
Using the radial reduction, the Apple laptop executed the extraction in a comparable time of 94ms per image and completed the training phase on average in less than a second (0.89 s), achieving negligible test time.

\subsection{Results on \PRISMDB}
\label{ssec:resPrismDB}

\begin{figure}
    \centering
    \includegraphics[width=0.85\linewidth]{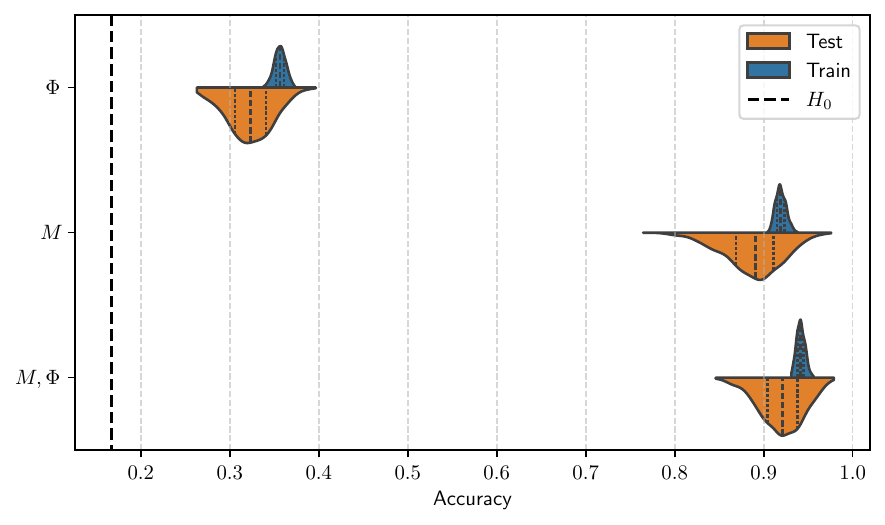}
    \caption{
        Distribution of test and training \textit{accuracies} over $N_s=1000$ random splits. The usage of all the features is compared with the sole Phase ($\Phi$) and the sole Magnitude ($M$).
        The dashed black line represents the Null hypothesis ($H_0$) where the class is randomly guessed.
    }
    \label{fig:LDA_ablation}
\end{figure}

We start presenting the results of the training/test statistics over $N_s = 1000$ splits of the datasets while preserving prompt separation, meaning that a couple of short-long prompts either appear in training or test set, but not both, actually ensuring correct prompt generalisation.
In particular, with average \textit{accuracy} being $92.04 \pm 2.43$ ($5$--$95\%$ being $87.88$--$95.85$, best results $97.89\%$), and comparable aggregated \fscore (up to 0.1\%), the method proved being robust and informative.Figure~\ref{fig:LDA_ablation} reports the distribution of the \textit{accuracies} within the $N_s$ random splits, comparing the results obtained using both Magnitude and Phase ($M, \Phi$) with the sole Magnitude ($M$), the sole Phase ($\Phi$), and the random guess ($H_0$). It can be seen that, despite the phase being insufficient on its own to train a discriminator, combined features still offer a sensible advantage over the sole magnitude. The adoption of phase significantly increases ($p\ll10^{-10}$ under paired $t$-test) the mean \textit{accuracy} by 3.23\%, while also reducing the standard deviation by $25.64\%$ --a marked and consistent improvement.

\begin{figure}[t]
    \centering
    \includegraphics[width=\linewidth]{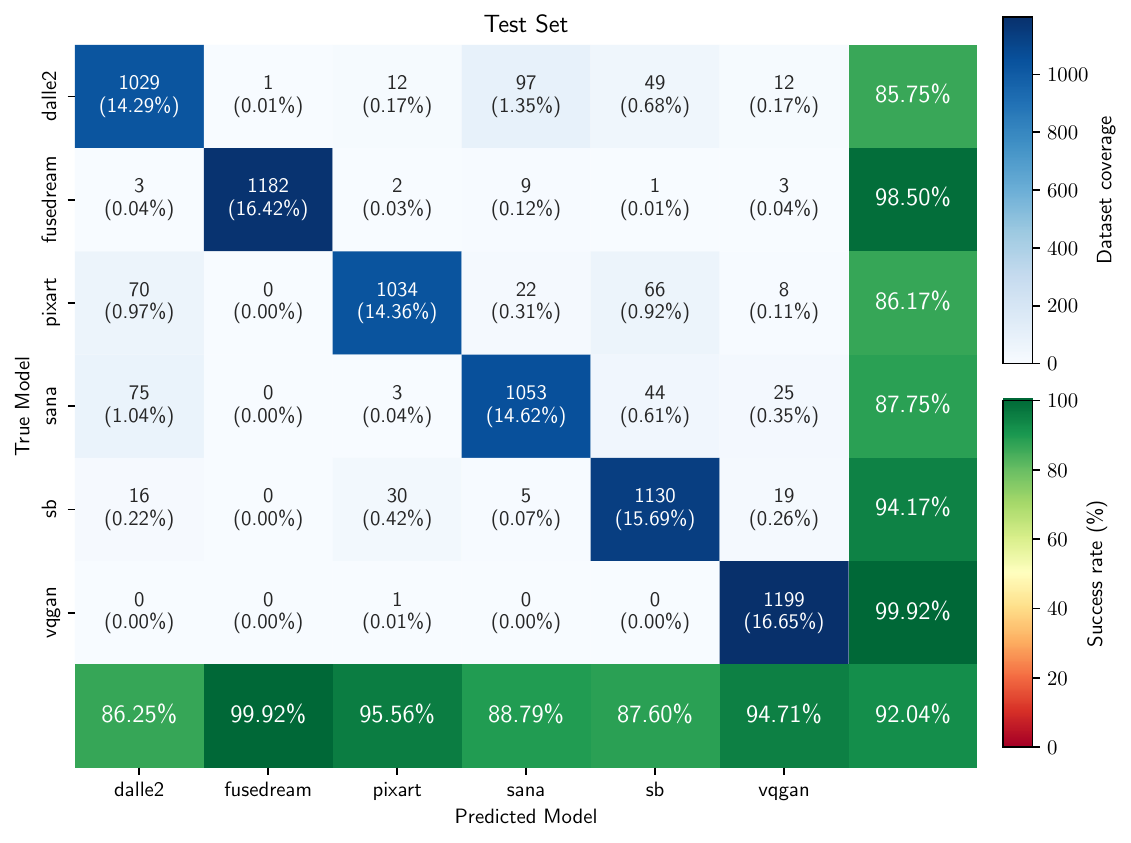}
    \caption{
        Confusion matrix of model predictions on the \PRISMDB test set. The last (external) row and column show model-wise \textit{precision} and \textit{recall} respectively, while their intersection represents the global \textit{accuracy}.}
    \label{fig:NORED-LDA-hemmedconfusion}
\end{figure}

Upon selecting the split achieving the closest results (min sum of squared errors) to the average \textit{accuracy}/\fscore, \ie the reference split, we present the corresponding confusion matrix in Figure~\ref{fig:NORED-LDA-hemmedconfusion}, highlighting an overall \textit{accuracy} of $92.04\%$ (in line with the mean \textit{accuracy}). 
Out of the considered datasets, \FUSEDREAM receives the most accurate attributions, almost reaching perfection with a \fscore of 0.99 ($1182$ correct, 18 false negatives, 1 false positive), followed closely by \VQGAN at $0.97\%$ (suffering from a weaker \textit{precision} despite the high \textit{recall} of a single false negative).
These results show that the proposed approach successfully learned highly discriminative features for these models.
\PIXART and \SBnv achieve strong performances as well, securing both a \fscore of 0.91, the first due to high \textit{precision} (95.56\%) and the latter due to high \textit{recall} (94.17\%).
Finally, \SANA and \DALLE offer the most challenging models to discriminate, jeopardising the overall score with less brilliant statistics, actually achieving balanced {\fscore}s of 0.88 and 0.86, respectively.
The \textit{precision} and \textit{recall} scores --corresponding respectively to the bottom row and right-most column of Figure~\ref{fig:NORED-LDA-hemmedconfusion}-- are however consistently high across all models, with no particular criticalities.
The confusion patterns are minimal (575 misclassifications out of 7200), with most occurring between \DALLE and \SANA which result mildly blended, accounting for 172 out of the total 575 misclassifications ($\sim 30\%$, distributed as $97$ and 75, respectively).
\SBnv is also frequently misclassified (27.65\% of the errors) in place of \PIXART, \DALLE, or \SANA, probably being ascribable to the shared architecture all these models are built onto (cf.\ Appendix~\ref{app:comparative_charts} -- Table~\ref{tab:our-dataset}).

\subsection{LDA Results on Reference Datasets}
\label{ssec:res-other-datasets}
\begin{figure}[t]
\footnotesize

    \centering
    \includegraphics[width=.85\columnwidth]{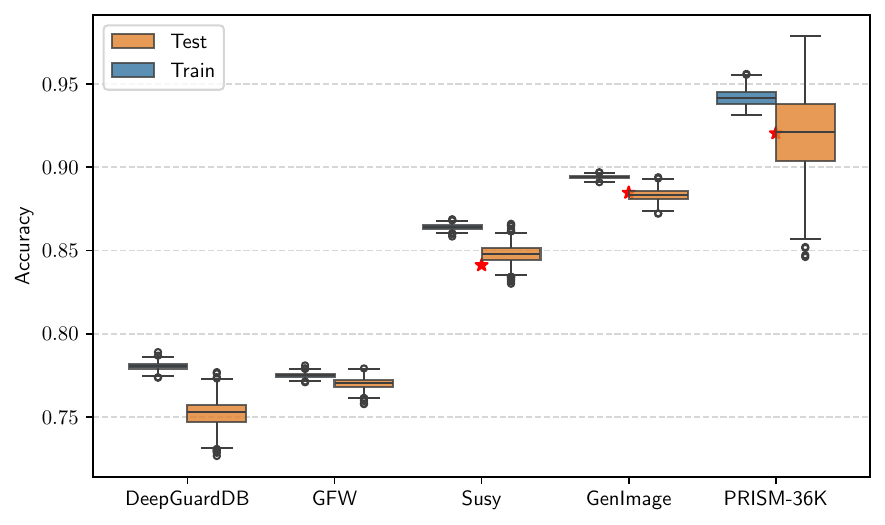}
    \caption{
        Performances boxplots (\textit{accuracy}) on the five different datasets.
        \fscore, \textit{precision}, and \textit{recall} are nearly indistinguishable.
        Red star reports the values against the published split (cf.\ Table~\ref{tab:model_attribution}).
    }
    \label{fig:boxplot}
\end{figure}

After evaluating our approach on \PRISMDB, we applied it to the other datasets presented in Section~\ref{ssec:other-dataset}, obtaining both the results shown in Table~\ref{tab:model_attribution} and in the boxplot represented in Figure~\ref{fig:boxplot}. While the latter shows the distribution of the results obtained on $N_s = 1000$ splits of each dataset, Table~\ref{tab:model_attribution} compares the results achieved on the average split of our dataset with the results obtained on the literature training-test splits for the \SUSY and \GENIMAGE datasets and on random splits for the remaining datasets.

\begin{table}[t]
    \footnotesize
    \centering
    \setlength{\tabcolsep}{3pt} \caption{Model attribution results.}
    \label{tab:model_attribution}
    \scalebox{.9}{\begin{tabular}{|c|c|c|c|c|}
    \hline
    \textbf{Dataset} & \textbf{Models} & \textbf{Train/Test} & \textbf{Accuracy} & \textbf{\fscore} \\
    \hline
    \PRISMDB & 6 & 28.8k/7.2k & 92.04 & 92.02 \\
    \hline
    \GENIMAGE & 8 & 28k/7k & 88.47 & 88.25 \\
    \hline
    \SUSY & 6 & 14.5k/5.6k & 84.12 & 84.00 \\
    \hline
    \GFW & 4 & 60.8k/15.2k$^\dagger$ & 77.07 & 76.55 \\
    \hline
    \DEEPGUARD & 5 & 10.4k/2.6k$^\dagger$ & 76.31 & 75.46 \\
    \hline
    \multicolumn{5}{l}{$^\dagger$ Random 80/20\% split. Others use published splits.}
    \end{tabular}}
\end{table}

In detail, Table~\ref{tab:model_attribution} shows that \PRISMDB is the most recognisable dataset, achieving the highest \textit{accuracy} of $92.04\%$ and \fscore of $92.02\%$; this is not surprising, considering that, conversely to the literature dataset, we have built it specifically to study the model attribution task. In contrast, \DEEPGUARD and \GFW represent the most challenging scenarios, with \textit{accuracies} of $76.31\%$ and $77.07\%$ respectively, and corresponding {\fscore}s of $75.46\%$ and $76.55\%$. The intermediate performance of \GENIMAGE ($88.47\%$ \textit{accuracy}, $88.25\%$ \fscore) and \SUSY ($84.12\%$ \textit{accuracy}, $84.00\%$ \fscore) demonstrates that dataset composition and the specific generative models included significantly influence attribution capability.

The boxplot in Figure \ref{fig:boxplot} reveals significant performance variations across the five datasets. Our dataset consistently performs better, showing \textit{accuracies} concentrated around $90.40$--$93.78\%$, indicating excellent classification \textit{accuracy}; however, major variance is obtained in Test \textit{accuracy}, actually suggesting some prompts might be correlated, further enhancing performances when considered split within training and test set. \GENIMAGE and \SUSY datasets show moderate performance with \textit{accuracies} ranging around $88.10$--$88.59\%$ and $84.45$--$85.13\%$ respectively, while \GFW and \DEEPGUARD perform considerably lower, with results localising around $76.80$--$77.24\%$ and $74.69$--$75.73\%$. The lower performance on \DEEPGUARD can likely be attributed to the presence of lossy compression formats (JPEG, WebP) degrading the frequency-domain signatures.

Finally, results presented in Table \ref{tab:real_vs_fake} show the scores of our approach when used for distinguishing between real and synthetic images across the four established reference datasets. The results, obtained without retraining for the specific task, demonstrate substantial variability in classification performance across datasets. \textit{Accuracy} is overall higher w.r.t.\ the model attribution task, further suggesting the model attribution task being a generalisation of the \textit{real} vs \textit{fake} one.
More in detail, \GENIMAGE achieved the highest overall performance with an \textit{accuracy} (and \fscore) of $95.0\%$, better than the performance achieved in the dataset reference paper~\cite{zhu2023genimage}.
Similarly, our approach performs well on \SUSY, yielding the same performance as per \GENIMAGE ($95.0\%$).
In contrast, as before, \GFW and \DEEPGUARD exhibited more modest performance levels, with \textit{accuracies} of $82.94\%$ and $81.62\%$, and {\fscore}s of $83.10\%$ and $81.41\%$ respectively.

\begin{table}[t]
    \centering
    \setlength{\tabcolsep}{2.8pt} \caption{Results achieved on the binary classification task of \textit{real} vs \textit{fake} image detection. }
\label{tab:real_vs_fake}
    \scalebox{.75}{
    \begin{tabular}{|c|c|c|c|c|}
    \hline
    \textbf{Dataset} & \textbf{Paper} & \textbf{Train/Test} & \textbf{Accuracy} & \textbf{\fscore} \\
    \hline
    \multirow{2}{*}{\GENIMAGE} & \cite{zhu2023genimage} & 28k/7k & 82.20 & - \\
    \cline{2-5}
    & Ours & 28k/7k & 95.06 & 95.05 \\
    \hline
    \SUSY & Ours & 14.5k/5.6k & 95.01 & 95.03 \\
    \hline
    \GFW & Ours & 60.8k/15.2k$^\dagger$ & 82.94 & 83.10 \\
    \hline
    \multirow{2}{*}{\DEEPGUARD} & \cite{namani2025deepguard} & 10k/3k & 99.87 & - \\
    \cline{2-5}
    & Ours & 10.4k/2.6k$^\dagger$ & 81.62 & 81.41 \\
    \hline
    \multicolumn{5}{l}{$^\dagger$ Random 80/20\% split. Others use published splits.}
    \end{tabular}}
\end{table}

\section{Discussion}\label{sec:discussion}
Our experiments confirm the effectiveness of frequency-domain fingerprinting for model attribution of AI-generated images in black-box conditions.
The combination of magnitude and phase embeddings enables the extraction of structured spectral cues that are distinctive across generative models.
The LDA-based classification pipeline achieves a strong average \textit{accuracy} of 92.04\% over 1000 random splits on \PRISMDB, peaking at 97.89\% in favourable cases.
Certain models exhibit unique and stable spectral signatures.
For instance, \FUSEDREAM and \VQGAN consistently produce fingerprints that are highly separable, with \fscore values of $ 99\%$ and $97\%$ respectively, suggesting the presence of stable and model-specific spectral biases likely rooted in the architectures or training procedures~\cite{li2025improving, giorgi2025human}. Conversely, models built on similar diffusion frameworks --such as \SBnv and \DALLE-- yield more overlapping spectral signatures, with lower attribution performances ($91\%$ to $86\%$).

The characteristics of the evaluation datasets also play a decisive role.
Our controlled \PRISMDB achieves the highest attribution scores, benefiting from a balanced and clean distribution of prompt-image pairs.
In contrast, public datasets introduce challenges stemming from heterogeneity in data curation.
\GENIMAGE performs robustly (88.47\% \textit{accuracy}), but suffers from severe class imbalance.
\SUSY remains moderately effective (84.12\%), while \GFW and \DEEPGUARD are the most challenging datasets, with an \textit{accuracy} of 77.07\% and 76.31\%, mainly ascribable to mixed resolutions, colour spaces, and lossy compression formats (JPEG, WebP).
These artefacts substantially degrade frequency-domain fidelity, making our approach to attribution more difficult. Nonetheless, we achieve the best results on \GFW and a competitive performance on \DEEPGUARD. This variability highlights the need for reproducible, standardised benchmarks for multi-class model attribution. Moreover, while most of the prior literature focuses on binary \textit{real} vs \textit{fake} detection~\cite{tan2024frequency}, our results confirm that full model attribution is a significantly harder task. For example, our method achieves 95.06\% \textit{accuracy} on the binary classification task for \GENIMAGE, outperforming~\cite{zhu2023genimage} by a large margin (95.06\% vs 82.20\%), while performing at 88.47\% on full attribution. This supports the conclusion that model attribution generalises and extends beyond the binary setting.

Finally, we note that model attribution may be impacted by semantic mismatches between prompts and generated content. Prior work~\cite{ricco2025hallucination} shows that generative artefacts and hallucinations can introduce spectral anomalies. These perturbations—whether due to prompt ambiguity or instability in the generative process—may reduce attribution \textit{accuracy} by introducing noise in the extracted frequency representation.

In summary, our findings validate the utility of frequency-based fingerprinting for model attribution and show that linear classifiers, such as LDA, represent an effective solution in this setting.
These results encourage further development of scalable, interpretable, and reproducible attribution pipelines that operate under realistic, black-box conditions.

\section{Conclusions} \label{sec:conclusions}
We proposed \PRISM, a novel methodology for fingerprinting AI-generated images based on frequency-domain representations. Our approach generalises the conventional \textit{real} vs \textit{fake} detection task into the more informative, yet challenging, model attribution problem. By extracting features from both the magnitude and the phase of the radially-reduced DFT, 
we manage to accurately distinguish generative sources via Linear Discriminant Analysis only relying on frequency-domain patterns.

To validate our methodology, we constructed \PRISMDB, a controlled dataset comprising 36,000 images from six prominent text-to-image models, each sampled with the same set of 40 prompts. In addition, we evaluated our approach on four public benchmark datasets to assess its robustness under real-world conditions. On \PRISMDB, our method achieves an average attribution \textit{accuracy} of 92.0\%, with performance peaking at 97.9\% in favourable splits. On benchmark datasets we obtain competitive results, reaching up to 88.5\% \textit{accuracy} on \GENIMAGE for the attribution task, and outclassing the reference accuracy in \textit{real} vs \textit{fake} classification (95.1\% against 82.2\% from the literature), hence demonstrating resilience even under adverse compression and class imbalance.

These results demonstrate the viability of frequency-based fingerprinting for model attribution, confirming that meaningful spectral artefacts are embedded in AI-generated content.
However, several research directions remain open. In future work, we plan to explore unsupervised clustering to extend applicability in low-supervision settings, experiment with alternative dimensionality reduction 
to detect embedded structural relationships in spectral features, and investigate their interpretability to better understand how they relate to generative model behaviour.

\bibliography{aaai2026}

\newpage

\appendix

\section{Comparative Charts}
\label{app:comparative_charts}

The interested reader can find a comparison chart of the various works involving multi-class attribution of AI-generated images in Table~\ref{tab:fingerprinting-comparison}.

\begin{table*}[hpt]
\centering
\caption{Overview of the different approaches to multi-class fingerprinting AI-generated images.}
\setlength{\tabcolsep}{2.9pt}
\label{tab:fingerprinting-comparison}
\scalebox{.8}{\begin{tabular}{|c|c|c|c|c|p{3cm}|}
\hline
\textbf{Paper} & \textbf{Task} & \textbf{Feature Extraction} & \textbf{Model} & \textbf{Learning} & \textbf{Dataset}\\

\hline

\multirow{2}{*}{\makecell{DE-FAKE \\ \cite{sha2023fake}}}
& \multirow{2}{*}{\textit{Real} vs \textit{fake} detection} 
& \multirow{2}{*}{Average Fourier Transform} 
& \multirow{2}{*}{CNN} 
& \multirow{2}{*}{Supervised}
& \texttt{MSCOCO} \\
& & & & & \texttt{Flickr30k} \\

\hline
\multirow{3}{*}{\makecell{AI Attribution\\ \cite{wang2023did}}} & \multirow{3}{*}{Origin Attribution} & \multirow{3}{*}{Reverse engineering} & \multirow{3}{*}{Optimisation Algorithm} & \multirow{3}{*}{Supervised}
& \texttt{CIFAR-10} \\
& & & & & \texttt{ImageNet} \\
& & & & & \texttt{CUB-200-2011} \\

\hline
\multirow{2}{*}{\makecell{Whodunit\\\cite{wissmann2024whodunit}}} & \multirow{2}{*}{Image Fingerprinting} & Discrete Cosine Transform & \multirow{2}{*}{ResNet50} & \multirow{2}{*}{Supervised} & \texttt{CelebA} \\
\cline{3-3} & & Power Spectral Density & & & Diffusion Models$^\star$ \\
\hline

\multirow{3}{*}{\makecell{Handcrafted Filters\\\cite{li2024handcrafted}}} & \multirow{3}{*}{Origin Attribution} & \multirow{3}{*}{Multi-Directional High-Pass Filters} & \multirow{3}{*}{ResNet50} & \multirow{3}{*}{Supervised} & \texttt{GAN} \\
& & & & & Diffusion Models$^\star$ \\
& & & & & \texttt{ImageNet} \\
\hline

\multirow{2}{*}{\makecell{Forensics Approach\\\cite{Jiang2024}}} & \multirow{2}{*}{Origin Attribution} & \multirow{2}{*}{CLIP, HR-NET} & \multirow{2}{*}{Supervised Contrastative Loss} & \multirow{2}{*}{Supervised} & Diffusion Models$^\star$ \\
& & & & & \texttt{MS-COCO} \\

\hline
\multirow{2}{*}{\makecell{FIDAVL\\\cite{keita2025fidavl}}} & \multirow{2}{*}{Origin Attribution} & \multirow{2}{*}{Image Encoder} & \multirow{2}{*}{Q-Former} & \multirow{2}{*}{Supervised} & \texttt{GAN} \\
& & & & & Diffusion Models$^\star$ \\
\hline

\makecell{DeepGuard\\\cite{namani2025deepguard}} & \multirow{1}{*}{Origin Attribution} & \multirow{1}{*}{Binary Classifier} & \multirow{1}{*}{Ensemble Learning} & \multirow{1}{*}{Supervised}
& \DEEPGUARD \\

\hline
\multirow{2}{*}{\makecell{Detect Origin\\\cite{xu2025detecting}}} & \multirow{2}{*}{Origin Attribution} & \multirow{2}{*}{Style features (Gram matrices), RGB} & \multirow{2}{*}{EfficientFormer} & \multirow{2}{*}{Supervised} & Diffusion Models$^\star$ \\
& & & & & \texttt{MS-COCO} \\
\hline

\multirow{5}{*}{\makecell{\textbf{\PRISM}\\ (\textit{Ours})}} 
& \multirow{5}{*}{{Image Fingerprinting}} 
& \multirow{5}{*}{{Radial DFT}} 
& {K-Means} 
& \multirow{3}{*}{{Unsupervised}}
& {\PRISMDB} \\
\cline{4-4}
& & & {Fuzzy-C-Means} & & \GFW \\
\cline{4-4}
& & & {GMM} & & \DEEPGUARD \\
\cline{4-5}
& & & \multirow{2}{*}{{LDA}} & \multirow{2}{*}{{Supervised}} & \SUSY \\
& & & & & \GENIMAGE\\

\hline

\end{tabular}}

\medskip
\scalebox{.8}{$^\star$Diffusion Models denotes a subset of \texttt{LDM}, \texttt{Glide}, \texttt{ADM}, \SB[1.4], \texttt{VQDM}, \DALLE, and \texttt{Midjourney}. Details are in the cited papers.}
\end{table*}

Also, Table~\ref{tab:different-dataset} present a comparison chart of the literature datasets of generated image compared to our proposed \PRISMDB.

\begin{table*}[hpt]
\centering
\footnotesize
\caption{Comparison of our dataset with other reference datasets.}
\label{tab:different-dataset}
\renewcommand{\arraystretch}{1.3} \setlength{\tabcolsep}{3pt} \scalebox{.8}{\begin{tabular}{|c|c|c|c|c|c|c|c|}
\hline
    \multirow{2}{*}{\textbf{Reference}} & 
    \multirow{2}{*}{\textbf{Dimension}} &
    \multirow{2}{*}{\textbf{Format}} & 
    \textbf{Ref.} & 
    \multicolumn{4}{c|}{\textbf{Models}} \\
\cline{5-8}
    & & & \textbf{Split} & \textbf{Name} & \textbf{Resolution} & \textbf{Dimension} & \textbf{Classes} \\
\hline

\makecell{\GFW\\\cite{borji2022generated}}
    & $76,000$
    & $.jpeg$
    & $\times$
    & \makecell{\DALLE \\ 
                \texttt{Midjourney} \\ 
                \SB[1.4] \\ 
                Real Faces$^*$}
    & \makecell{$100 \times 100$, $512 \times 512$ \\ 
                $100 \times 100$, various  \\ 
                $100 \times 100$, $512 \times 512$ \\ 
                $100 \times 100$}
    & \makecell{$1,109$ \\ 
                $18,609$ \\ 
                $26,282$ \\ 
                $30,000$}
    & \makecell{Faces, Images \\ 
                Faces, Images \\ 
                Faces, Images \\ 
                Faces} \\

\hline

\makecell{\GENIMAGE\\ \cite{zhu2023genimage}\\ (\textit{Tiny}) }
    & $35,000$
    & \makecell{$.jpeg$\\$.png$}
    & $\checkmark$
    & \makecell{
                Nature$^*$ \\
                \texttt{ADM} \\       
                \texttt{BiGGAN} \\
                \texttt{GLIDE} \\
                \texttt{Midjourney} \\
                \SB[1.5] \\ 
                \texttt{VQDM} \\
                \texttt{Wukong} 
                }
    & \makecell{various \\
                $256 \times 256$  \\
                $128 \times 128$  \\
                $256 \times 256$  \\
                $1024 \times 1024$ \\
                $512 \times 512$ \\ 
                $256 \times 256$  \\ 
                $512 \times 512$  \\ 
                }
    & \makecell{$17,500$ \\
                $2,500$ \\ 
                $2,500$ \\ 
                $2,500$ \\ 
                $2,500$ \\  
                $2,500$ \\ 
                $2,500$ \\ 
                $2,500$ \\ 
                }
    & \makecell{\textit{Real} \\    
                \textit{Fake} \\ 
                \textit{Fake} \\ 
                \textit{Fake} \\ 
                \textit{Fake} \\  
                \textit{Fake} \\
                \textit{Fake} \\ 
                \textit{Fake} \\
                } \\
\hline

\makecell{\SUSY\\\cite{bernabeu2024present}}
    & $20,004$
    & \makecell{$.jpeg$\\$.png$}
    & $\checkmark$
    & \makecell{\texttt{COCO} \\
                \texttt{DALL-E 3} \\
                \SB[1.X] \\
                \texttt{SDXL} \\ 
                \texttt{Midjourney-tti} \\   
                \texttt{Midjourney-img} }
    & \makecell{various \\
                various \\
                $512 \times 512$ \\
                various \\
                $1024 \times 1024$ \\ 
                $1024 \times 1024$}
    & \makecell{$4,201$ \\ 
                $1,317$ \\ 
                $4,200$ \\ 
                $4,201$ \\ 
                $3,624$ \\ 
                $2,461$}
    & \makecell{\textit{Real}$^*$ \\    
                \textit{Fake} \\ 
                \textit{Fake} \\ 
                \textit{Fake} \\  
                \textit{Fake} \\
                \textit{Fake}} \\
\hline
\makecell{\DEEPGUARD\\\cite{namani2025deepguard}}
    & $13,000$
    & \makecell{$.png$\\$.jpeg$\\$.webp$\\(lossy)}
    & $\times$
    & \makecell{Real \\
                \texttt{DALL-E 3} \\ 
                \SB[3.0] \\ 
                \texttt{GLIDE} \\ 
                \texttt{IMAGEN} }
    & \makecell{$512\times 512$} 
    & \makecell{
                $6,500$ \\
                $2,150$ \\ 
                $2,675$ \\ 
                $500$ \\ 
                $1,175$}
    & \makecell{\textit{Real} \\ 
                \textit{Fake} \\ 
                \textit{Fake} \\ 
                \textit{Fake} \\ 
                \textit{Fake}} \\
\hline
\makecell{\textbf{\PRISMDB}\\(\textit{Ours})}
    & $36,000$
    & $.png$
    & $\checkmark$
    & \makecell{\DALLE \\
                \FUSEDREAM \\
                \PIXART  \\
                \SANA  \\
                \SB  \\
                \VQGAN  
                }
    & $512 \times 512$
    & \makecell{$6,000$ \\ 
                $6,000$ \\ 
                $6,000$ \\ 
                $6,000$ \\ 
                $6,000$ \\ 
                $6,000$}
    & \makecell{20 long prompts\\\& \\20 short prompts}\\
\hline

\end{tabular}}
\end{table*}

Finally, Table~\ref{tab:our-dataset} presents the details of the six T2I models used to generate the \PRISMDB.

\begin{table*}[hpt]
\centering
\caption{Details of the T2I models used to generate the \PRISMDB dataset.}
\label{tab:our-dataset}
\scalebox{.7}{\begin{tabular}{|c|c|c|cc|c|}
\hline
\multirow{2}{*}{\textbf{Model}} & \multirow{2}{*}{\textbf{Reference}} & \multirow{2}{*}{\textbf{Resolution}} & \multicolumn{2}{c|}{\textbf{Architecture}} & \multirow{2}{*}{\textbf{Openness}} \\
\cline{4-5}
 & & & \textbf{GAN+CLIP} & \textbf{Transformers+Diffusion} &  \\
\hline
\VQGAN & \makecell{\cite{esser2021taming} \\ \cite{li2022clip}}
  & Custom$^\star$
  & $\checkmark$ &  
  & Partially Open \\
  
\hline
\SB[1.4] & \cite{Rombach_2022_CVPR} 
  & $512 \times 512$ 
  & & $\checkmark$  
  & Fully Open \\
\hline
\SANA & \cite{xie2024sana} 
  & Custom$^\star$
  & & $\checkmark$  
  & Partially Open \\
\hline
\PIXART & \cite{chen2024pixart} 
  & Custom$^\star$ 
  & & $\checkmark$  
  & Partially Open \\
\hline
\FUSEDREAM & \cite{liu2021fusedream}
  & $256 \times 256$, $512 \times 512$  
  & $\checkmark$ &   
  & Partially Open \\
\hline
\DALLE & \cite{ramesh2022hierarchical}
  & $256 \times 256$, $512 \times 512$, $1024 \times 1024$    
  & & $\checkmark$    
  &  Closed \\
\hline
\end{tabular}}

\medskip
\scalebox{.87}{$^\star$\textit{Custom} refers to the possibility for the user to decide the resolution.}
\end{table*}

\section{Prompts} \label{app:labels}

For the sake of completeness, we report in what follows the short and long prompts used to generate the \PRISMDB dataset.

\begin{tcolorbox}[coltext=black, width=\linewidth, halign=justify, breakable]
\textit{Short prompts:}
\begin{enumerate}
    \item A cat lying on the table
    \item A dog running on grass
    \item A man walking on sand
    \item An astronaut flying into space
    \item A girl swimming in the sea
    \item An apple in a fruit bowl
    \item A starred night in the dark
    \item A sailboat in the stormy sea
    \item A bird perched on a tree
    \item A scene from War and Peace
    \item A butterfly landing on a flower
    \item A child playing in the snow
    \item A chef cooking in a kitchen
    \item A car driving through mountains
    \item A painter working on a canvas
    \item A wave crashing against rocks
    \item A robot building a structure
    \item A fox hiding in bushes
    \item A balloon floating in the sky
    \item A writer typing on a laptop
\end{enumerate}
\end{tcolorbox}

\begin{tcolorbox}[coltext=black, width=\linewidth, halign=justify, breakable]
\textit{Long prompts:}
\begin{enumerate}\setcounter{enumi}{20}
    \item A cat lying on the table while its owner throws it a yellow and red ball to play with
    \item A dog running across the grass towards a child waiting for him with open arms
    \item A man walking on the beach with his son on the shore of a stormy sea with a purple kite
    \item An astronaut flying in space observing Planet Earth from black space in his white suit
    \item A girl swimming in the sea with curly blond hair next to a white sailboat with a black hull
    \item An apple in a fruit bowl during a huge sandstorm in the red Sahara desert near a waterhole
    \item A starred night in the dark as a shining comet passes by with a vortex of lightning stars
    \item A sailboat in the stormy sea under a black hole with flying fairies playing classical music
    \item A bird perched on a tree with code flowing through its branches with a clock showing 12 am
    \item A scene from War and Peace reinterpreted in a futuristic landscape with robots instead of humans
    \item A butterfly landing on a flower while a photographer tries to capture the moment
    \item A child playing in the snow with red mittens while building an ice castle that glows with blue light
    \item A chef cooking in a kitchen filled with steam as magical ingredients float in the air
    \item A car driving through mountains on a winding road with colorful autumn leaves swirling around
    \item A painter working on a canvas while the subjects of the painting step into the real world
    \item A wave crashing against rocks revealing an ancient underwater city with merfolk observers
    \item A robot building a structure made of translucent crystals on the surface of Mars
    \item A fox hiding in bushes with a pocket watch around its neck as it waits for the perfect moment
    \item A balloon floating in the sky carrying a tiny house with a chimney that releases rainbow smoke
    \item A writer typing on a laptop beside a window showing different dimensions with each keystroke
\end{enumerate}
\end{tcolorbox}

\section{DFT and Radial Aggregation} \label{app:DFT}

In what follows, we discuss in detail the feature extraction procedure adopted in Section~\ref{ssec:rFFT}.

Given an input image with resolution $n_x \times n_y$ and pixel values in $\NN_{256}$, we consider each colour channel independently.
Let $I \in \NN_{256}^{n_y \times n_x}$ denote one such channel (in the case of our \PRISMDB, $n_x = n_y = 512$).
The two-dimensional Discrete Fourier Transform (DFT) of $I$ is the complex-valued matrix $\widehat{I} \in \CC^{n_y \times n_x}$ defined by:
\begin{equation}
    \widehat{I}_{u,v} = \sum_{x=0}^{n_x-1} \sum_{y=0}^{n_y-1} I_{x,y} \, e^{-2\pi i \left( \frac{ux}{n_x} + \frac{vy}{n_y} \right)}\ ,
\end{equation}
where each entry $\widehat{I}_{u,v}$ encodes the amplitude and phase of a spatial frequency component as the magnitude and angle of the complex value.

To facilitate interpretation, we perform a standard frequency centralisation by shifting the zero-frequency component (DC component) to the centre of the matrix. This is done via a two-dimensional circular shift:
\begin{equation}
    \tilde{I}_{u,v} = \widehat{I}_{(u + \sfrac{n_x}2) \bmod n_x, (v + \sfrac{n_y}2) \bmod n_y},
\end{equation}
so that lower frequencies are positioned at the centre of the spectrum, with higher frequencies distributed symmetrically outward.

From the centralised spectrum $\tilde{I}$, we extract:
\begin{equation}
    \overline{M}_{u,v} = \log\left( \left|\tilde{I}_{u,v}\right| + 1 \right)
    \ ,\quad
    \overline{\Phi}_{u,v} = \arg(\tilde{I}_{u,v})\ ,
\end{equation}
representing the log-scaled magnitude and phase at frequency $(u, v)$. The log-scaling improves numerical stability and compresses dynamic range.

\medskip

To obtain a compact and dimension-agnostic descriptor, we compute a radial DFT profile (rDFT). This aggregates spectral information across annular frequency regions. Specifically, for each frequency index $(u, v)$, we compute its Euclidean distance from the centre:
\begin{equation}
r(u,v) = \| (u, v) - \left(\sfrac{n_x}{2}, \sfrac{n_y}{2} \right) \|_2\ ,
\end{equation}
and assign it to one of $n_r = 64$ evenly spaced radial bins $\BBB_1, \dots, \BBB{n_r}$ covering the full frequency range.

Then, for each bin $i = 1, \dots, n_r$, we define the aggregated log-magnitude and phase as in \eqref{eq:M_i} and \eqref{eq:Phi_i} we here report for completeness:
\begin{align}
M_i &= \frac{1}{|\BBB_i|} \sum_{(u,v) \in \BBB_i} \overline{M}{u, v}, \\
\Phi_i &= \cos\left( \arg\left( \frac{1}{|\BBB_i|} \sum{(u,v) \in \BBB_i} e^{i \overline{\Phi}_{u,v}} \right)\right)\ ,
\end{align}
where $\arg(\cdot)$ denotes the complex argument and the outer cosine transform is used to identify $\pm\pi$ by collapsing opposite phases directions (\ie $0\pm\epsilon$ and $\pm\pi\mp\epsilon$) onto a consistent representation. This is especially justified by the Hermitian property of the DFT (due to the real-valued input), which ensures phase symmetry and supports this form of aggregation, yet potentially being subject to numerical issues $\epsilon \ne 0$.

The result is a fixed-size, rotation-invariant vector that captures the spectral distribution of both amplitude and phase across radial frequency bands. This transformation also promotes robustness to variations in image resolution and compression artefacts.

\end{document}